\documentclass{article}

\usepackage{microtype}
\usepackage{graphicx}
\usepackage{subcaption}
\usepackage{booktabs} 

\usepackage{hyperref}

\usepackage[preprint]{icml2026}

\usepackage{amsmath}
\usepackage{amssymb}
\usepackage{mathtools}
\usepackage{amsthm}

\usepackage[capitalize,noabbrev]{cleveref}

\theoremstyle{plain}

\theoremstyle{definition}

\theoremstyle{remark}

\usepackage[textsize=tiny]{todonotes}

\usepackage{amsmath,amsfonts,bm}

\def\eqref#1{equation~\ref{#1}}

\def\1{\bm{1}}

\DeclareMathAlphabet{\mathsfit}{\encodingdefault}{\sfdefault}{m}{sl}
\SetMathAlphabet{\mathsfit}{bold}{\encodingdefault}{\sfdefault}{bx}{n}

\usepackage{graphicx}

\usepackage{hyperref}
\hypersetup{colorlinks=true, urlcolor=blue}
\usepackage[most]{tcolorbox}

\usepackage{threeparttable}

\usepackage{cleveref}
\usepackage{url}
\usepackage{caption}
\usepackage{tabularx}
\usepackage{adjustbox}
\usepackage{booktabs}
\usepackage{makecell}
\usepackage{diagbox}
\usepackage{pifont}

\usepackage{graphicx}
\usepackage{subcaption}
\usepackage{wrapfig}

\usepackage{listings}

\usepackage{multirow}

\usepackage{amsthm}

\newcommand{\ie}{\textit{i}.\textit{e}.}
\newcommand{\eg}{\textit{e}.\textit{g}.}
\usepackage{cleveref}

\crefname{section}{Sec.}{Secs.}
\crefname{figure}{Fig.}{Figs.}
\crefname{table}{Tab.}{Tabs.}
\crefname{equation}{Eq.}{Eqs.}
\usepackage{multicol} 
\usepackage[table,dvipsnames,rgb]{xcolor}
\definecolor{darkgreen}{RGB}{0,100,0} 
\definecolor{colorpi0}{rgb}{0.85, 0.85, 1.0}
\definecolor{colorpi05}{rgb}{0.85, 1.0, 0.85}
\usepackage[utf8]{inputenc}
\usepackage[T1]{fontenc}
\usepackage{tcolorbox}
\usepackage{lipsum}
\newcommand{\deltaimp}[1]{\color{darkgreen}+#1}

\definecolor{myblue}{HTML}{ECF2F8}
\definecolor{mygreen}{HTML}{EEF8EC}
\icmltitlerunning{$\pi_\texttt{RL}$: Online RL Fine-tuning for Flow-based Vision-Language-Action Models}

\begin{document}

\twocolumn[{
\icmltitle{$\pi_\texttt{RL}$: Online RL Fine-tuning for Flow-based Vision-Language-Action Models}

\icmlsetsymbol{cor}{$\dagger$}
\vspace{-1em}
\begin{icmlauthorlist}
\icmlauthor{Kang Chen$^{2,6,*}$}{}
\icmlauthor{Zhihao Liu$^{3,6,*}$}{}
\icmlauthor{Tonghe Zhang$^{4,*,\sharp}$}{}
\icmlauthor{Zhen Guo$^{5}$}{}
\icmlauthor{Si Xu$^{5}$}{}
\icmlauthor{Hao Lin$^{5}$}{}

\icmlauthor{Hongzhi Zang$^{1}$}{}
\icmlauthor{Xiang Li$^{5}$}{}
\icmlauthor{Bingwen Wei$^{1}$}{}
\icmlauthor{Jiakai Zhou$^{1}$}{}
\icmlauthor{Quanlu Zhang$^{5}$}{}

\icmlauthor{Zhaofei Yu$^{2}$}{}
\icmlauthor{Guoliang Fan$^{3}$}{}
\icmlauthor{Tiejun Huang$^{2}$}{}
\icmlauthor{Yu Wang}{thu,cor}
\icmlauthor{Chao Yu}{thu,cor}
\end{icmlauthorlist}

\icmlaffiliation{thu}{Tsinghua University 
$^{2}$Peking University 
$^{3}$Institute of Automation, Chinese Academy of Sciences 
$^{4}$Carnegie Mellon University
$^{5}$Infinigence AI 
$^{6}$Zhongguancun Academy
$\sharp$ Work done at Tsinghua University
$*$ Equal Contributions
}

\icmlcorrespondingauthor{Yu Wang}{yu-wang@tsinghua.edu.cn}
\icmlcorrespondingauthor{Chao Yu}{zoeyuchao@gmail.com}

\icmlkeywords{Machine Learning, ICML}

\vspace{0.2em}

\begin{center}
  \href{https://github.com/RLinf/RLinf}{%
    \raisebox{-0.4ex}{\includegraphics[height=1.2em]{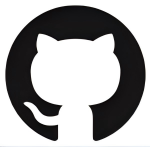}} \,https://github.com/RLinf/RLinf
  }
  \href{https://huggingface.co/RLinf}{%
    \raisebox{-0.4ex}{\includegraphics[height=1.2em]{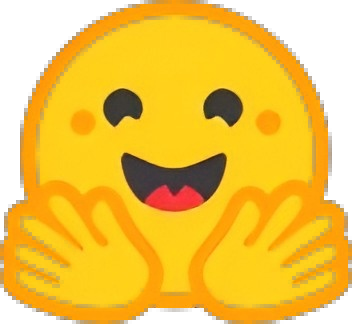}} \,https://huggingface.co/RLinf
  } 
\end{center}

\vspace{0.2em}

\begin{center}
    \includegraphics[width=1.0\textwidth]{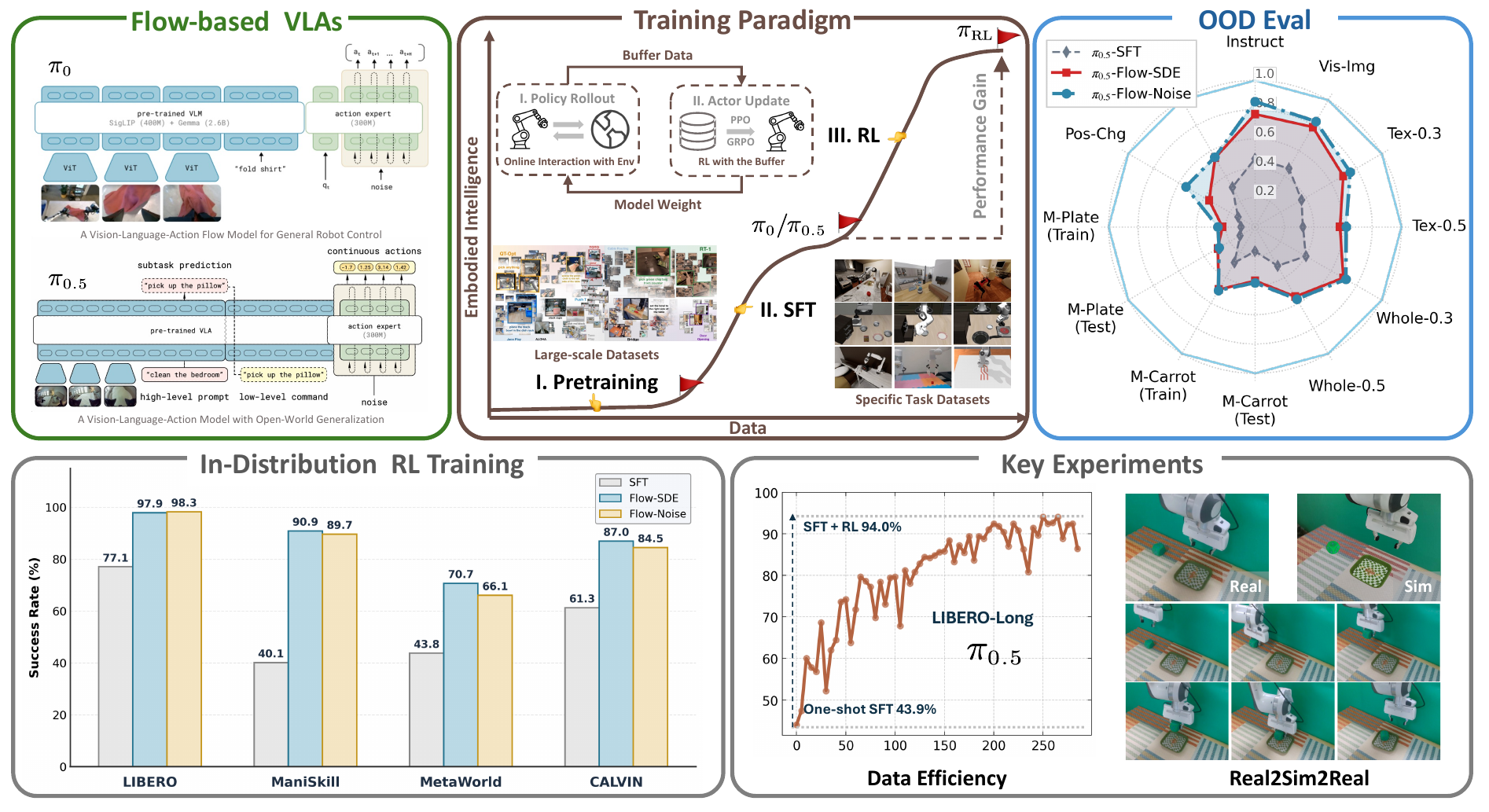}
    \captionof{figure}{
    $\pi_{\texttt{RL}}$: An online RL framework for flow-based VLAs. Incorporating two solutions, Flow-Noise and Flow-SDE, $\pi_{\texttt{RL}}$ enhances the performance and generalization of SFT-aligned models across extensive ID benchmarks and OOD settings. Refined with RL, few-shot SFT policies achieve performance comparable to full dataset baselines. 
    Additionally, we facilitate seamless zero-shot sim-to-real transfer by constructing a simulator with 3D Gaussian Splatting as the rendering engine to narrow the visual domain gap.
    \vspace{1em}
    }
    \label{fig:top_figure}
\end{center}
}]



\printAffiliationsAndNotice{}  
\begin{abstract}
Vision-Language-Action (VLA) models enable robots to understand and perform complex tasks from multimodal input. Although recent work explores using reinforcement learning (RL) to automate the laborious data collection process in scaling supervised fine-tuning (SFT), applying RL to large-scale  flow-based VLAs (\eg, $\pi_0$, $\pi_{0.5}$) remains challenging due to intractable action log-likelihoods raised from flow matching.
We address this challenge with $\pi_{\texttt{RL}}$, featuring two technical approaches:
(1) \textbf{Flow-Noise} models the denoising process as a discrete-time MDP with a learnable noise network for exact log-likelihood computation.
(2) \textbf{Flow-SDE} integrates denoising with agent-environment interaction, formulating a two-layer MDP that employs ODE-to-SDE conversion for efficient RL exploration.
We evaluate $\pi_{\texttt{RL}}$ across various benchmarks, with experiments demonstrating that RL yields significant performance improvements in both in-distribution and out-of-distribution settings.
\end{abstract}

\section{Introduction}
Vision-Language-Action (VLA) models \citep{din2025vision} have emerged as a leading solution for general-purpose robots, effectively bridging the gap between high-level multimodal reasoning and low-level physical control \citep{firoozi2025foundation}. Conditioned on sensor inputs and language commands, VLAs \citep{team2024octo,kim2024openvla,black2024pi_0,intelligence2025pi05} can translate abstract instructions into executable robotic actions, thereby enabling intuitive and flexible human-robot interaction.

The training methodology for VLAs follows the standard pre-training and supervised fine-tuning (SFT) paradigm as shown in \cref{fig:top_figure}. Building on the pretrained Vision-Language Model (VLM) \citep{touvron2023llama,beyer2024paligemma}, VLAs are fine-tuned on large-scale, heterogeneous human demonstration datasets \citep{o2024open,khazatsky2024droid}, followed by SFT on the target task to align their capabilities with the specific embodiment and environment. However, reliance on SFT introduces a critical challenge: curating large-scale, high-quality expert trajectories is both laborious and costly \citep{din2025vision}. Besides, models obtained via SFT tend to overfit to expert demonstrations \citep{liberoplus}, with their performance fundamentally constrained by the quality of expert demonstrations.

Recent efforts \citep{zang2025rlinf, li2025simplevla, tan2025riptvla, rl4vla} have explored expanding the VLA training process with reinforcement learning (RL), establishing a pre-training, SFT, and RL paradigm as shown in \cref{fig:top_figure}, allowing VLAs to improve their performance beyond expert demonstrations through environmental interaction and develop more generalizable policies.

However, these RL advances have been largely confined to autoregressive VLAs, featuring OpenVLA \citep{kim2024openvla} and OpenVLA-OFT \citep{kim2025openvlaoft}, which employ discrete action decoders that generate output in an autoregressive or parallel fashion. This stands in stark contrast to flow-based VLAs, exemplified by the $\pi$ series models, which generate actions through iterative refinement in flow matching \citep{lipman2022flow}, offering the advantages of generating action chunks in high-frequency and performing highly dexterous tasks \citep{black2024pi_0}. Consequently, previous VLA-RL algorithms are incompatible with flow-based VLAs, and the fundamental challenge lies in how to characterize a logarithmic likelihood \citep{hutchinson1989stochastic,chen2018neural} for the executed actions. 

In this paper, we introduce $\pi_{\texttt{RL}}$, a framework designed for fine-tuning flow-based VLAs with online RL algorithms. 
To address the intractable log-likelihood estimation problem in flow matching, we propose two solutions. 
\textbf{Flow-Noise} integrates a learnable noise network into the denoising process and models this stage as a discrete-time Markov decision process (MDP) for exact log-likelihood estimation. 
\textbf{Flow-SDE} converts the ordinary differential equation (ODE) denoising process into a stochastic differential equation (SDE) while maintaining equivalent marginal distributions for exploration, and builds a two-layer MDP that couples the denoising process with policy-environment interaction. 
Given the formulated MDP and the exact log-likelihood computation, $\pi_{\texttt{RL}}$ undergoes further optimization via the proximal policy optimization (PPO) \citep{schulman2017ppo}.

We conduct extensive experiments on various benchmarks to evaluate the effectiveness of $\pi_{\texttt{RL}}$ on $\pi_0$ \citep{black2024pi_0} and $\pi_{0.5}$ \citep{intelligence2025pi05} models. Across all benchmarks, the proposed framework consistently yields substantial performance gains over SFT baselines. Furthermore, out of distribution evaluations confirm that our model yields genuine policy enhancement rather than narrow overfitting on the target environment.


To sum up, our contributions are:
\begin{itemize}
    \item \textbf{RL for flow-based VLAs.} We introduce $\pi_{\texttt{RL}}$, an online RL fine-tuning framework with Flow-Noise and Flow-SDE formulations for flow-based VLAs. 

    \item \textbf{Superior Performance.} We demonstrate significant performance improvements and enhanced generalization of $\pi_{\texttt{RL}}$ across various benchmarks. 

    \item \textbf{Comprehensive Ablation.} We conduct thorough ablation studies, offering empirical insights to guide future RL research on flow-based VLAs.

    \item \textbf{Open-source Code and Models.} We release all codes to ensure reproducibility, hoping that our study helps to advance further research in this field.
\end{itemize}

\section{Related Work}
\subsection{Vision-Language-Action Models}
VLA models have recently achieved remarkable progress in robotics by integrating multimodal inputs to enable unified perception, reasoning, and control. This development has led to a series of architectures, including Octo \citep{team2024octo}, RT \citep{brohan2022rt}, OpenVLA, OpenVLA-OFT, $\pi_0$, $\pi_{0.5}$, and GR00T \citep{bjorck2025gr00t}. 


\subsection{Online RL Fine-tuning for VLA Models}
Recent research has increasingly focused on enhancing the performance and generalization of VLAs with online RL. For example, 
SimpleVLA-RL \citep{li2025simplevla}, building on the OpenVLA-OFT and GRPO, demonstrated that RL can improve long-horizon planning of VLA models under data scarcity. 
RL4VLA \citep{rl4vla} empirically evaluated PPO, GRPO, and direct preference optimization (DPO) \citep{rafailov2023dpo} with stage-based sparse rewards. 
RLinf-VLA \citep{yu2025rlinf, zang2025rlinf} provides a unified and efficient framework for scalable RL training of VLA models. These works demonstrate the effectiveness of RL fine-tuning VLA models.


\subsection{RL Fine-tuning for Flow Models}
Integrating RL with flow models is a promising way to transcend the limitations of imitation learning. To this end, 
Flow-GRPO \citep{liu2025flowgrpo} converts the deterministic ODE into an equivalent SDE to enable stochasticity exploration, a foundation upon which subsequent works like Mix-GRPO \citep{li2025mixgrpo} and TempFlow-GRPO \citep{he2025tempflow} further accelerate training through hybrid ODE-SDE rollouts. 
ReinFlow \citep{zhang2025reinflow} injects learnable noise into the flow path and transforms it into a discrete-time Markov process with a tractable likelihood for stable policy gradient updates. 
Flow policy optimization (FPO) \citep{mcallister2025fpo} reframes policy optimization as maximizing the advantage-weighted ratio of the conditional flow matching loss. 


\section{Preliminary}
\subsection{Problem Formulation}
\label{subsec:mdp}
We formulate the task as an MDP, defined by a tuple $\mathcal{M} = (\mathcal{S}, \mathcal{A}, P_0, P_{\text{ENV}}, R_{\text{ENV}},\gamma)$. The state $s_t \in \mathcal{S}$ is defined as the robot observation $\mathbf{o}_t$ and $P_0$ denotes the initial state distribution. Given the state, the flow policy predicts an action $a_t \sim \pi(\cdot |s_t) \in \mathcal{A}$, resulting in the state transition $s_{t+1} \sim P_{\text{ENV}}(\cdot|s_t,a_t)$ and a reward $R_{\text{ENV}}(s_t,a_t)$. The objective is to learn a policy $\pi_{\theta}$ that maximizes the expected $\gamma$-discounted return over a horizon of $T+1$:
\begin{equation}
    \mathcal{J}(\pi_\theta) = \mathbb{E}_{\pi_\theta,P_0} \left[ \sum_{t=0}^{T} \gamma^t R_{\text{ENV}}(s_t, a_t) \right].
\end{equation}

With the policy gradient surrogate \citep{williams1992simple}, the gradient of the return expectation can be approximated from sampled trajectories:
\begin{equation}
\label{equ:policy_opt}
    \nabla_\theta \mathcal{J}(\pi_\theta) = \mathbb{E}_{\pi_\theta,P_0} \left[ \sum_{t=0}^{T}\nabla_\theta \log \pi_\theta(a_t | s_t) A(s_t, a_t) \right].
\end{equation}
The advantage function, $A(s_t, a_t) = Q(s_t, a_t) - V(s_t)$, measures the relative merit of the action value $Q(s_t,a_t)$ over the state value $V(s_t)$, providing a low-variance signal for the policy update.

\subsection{Flow-based Vision-Language-Action Model}
\label{subsec:vla}
A flow-based VLA model $\pi_{\theta}$ is designed to map the observation $\mathbf{o}_t$ comprising RGB images, language tokens, and robot proprioception to a sequence of $H$ future actions $\mathbf{A}_t = [a_{t,0}, ..., a_{t,H-1}]$, formulated as $p(\mathbf{A}_t | \mathbf{o}_t)$. Within the model, the VLM extracts features from the visual and language inputs, while the flow matching expert is tasked with generating the actions. Specifically, the model learns a conditional vector field $\mathbf{v}_\theta$ that transforms a standard Gaussian noise distribution into the target action $\mathbf{A}_t$. This is achieved by minimizing the Conditional Flow Matching (CFM) loss, which aligns the predicted vector field $\mathbf{v}_\theta$ with the ground-truth vector field $\mathbf{u}$:
\begin{equation}
    \mathcal{L}_{\text{CFM}} = \mathbb{E}_{\tau,p(\mathbf{A}_t , \mathbf{o}_t), q(\mathbf{A}_t^\tau | \mathbf{A}_t)} \left[ \left\| \mathbf{v}_\theta(\mathbf{A}_t^\tau, \mathbf{o}_t) - \mathbf{u}(\mathbf{A}_t^\tau | \mathbf{A}_t) \right\|_2^2 \right].
\end{equation}
Here, the conditional probability path $q(\mathbf{A}_t^\tau | \mathbf{A}_t)$ generates a noisy action\footnote{$\mathbf{A}_{t}^{\tau}$ incorporates two temporal indices, $t$ denotes the discrete time step for environment interaction and $\tau$ represents the continuous time variable in flow matching.} $\mathbf{A}_t^\tau = \tau \mathbf{A}_t + (1-\tau)\epsilon$ from an action $\mathbf{A}_t$, random noise $\epsilon \sim \mathcal{N}(0, I)$, and a continuous time $\tau \in [0, 1]$ in flow matching. For this specific path, the corresponding ground-truth vector field is defined as $\mathbf{u}(\mathbf{A}_t^\tau | \mathbf{A}_t) = \mathbf{A}_t - \epsilon$. 

During the inference, the action sequence is generated by first sampling a noise vector $\mathbf{A}_t^0 \sim \mathcal{N}(0, I)$, which is further iteratively refined by integrating the learned vector field $\mathbf{v}_\theta$ over a fixed number of steps based on the forward Euler method: $\mathbf{A}_t^{\tau+\delta} = \mathbf{A}_t^\tau +  \mathbf{v}_\theta(\mathbf{A}_t^\tau, \mathbf{o}_t) \cdot \delta$.


\section{Methodology}
Existing VLA-RL approaches leverage base models such as OpenVLA for discrete actions and OpenVLA-OFT for continuous actions. To compute the action log-likelihood $\log \pi_\theta(a_t | s_t)$, discrete models \citep{rl4vla} apply softmax to the output logits, while continuous models \citep{li2025simplevla} treat the action as a Gaussian distribution, employing a prediction head to estimate the variance. As for the flow-based VLAs, directly computing the exact likelihood \citep{hutchinson1989stochastic} is inaccurate with few denoising steps. Moreover, the deterministic nature of its ODE sampling process precludes exploration, making its implementation within RL non-trivial. To this end, we propose Flow-Noise and Flow-SDE, two technical approaches that make flow-based VLAs amenable to RL.

\subsection{Flow-Noise}
Inspired by Reinflow \citep{zhang2025reinflow}, we incorporate a learnable noise network into the flow matching denoising process and solve the problem within the standard one-layer MDP framework detailed in \cref{subsec:mdp}. By modeling the denoising stage as a discrete MDP, we can directly compute the log-likelihood of the denoised sequence, enabling equivalent policy optimization via RL.

\subsubsection{Stochasticity Injection}
\label{subsubsec:learnable_noise}
\begin{figure}[!t]
    \centering
    \includegraphics[width=1.0\linewidth]{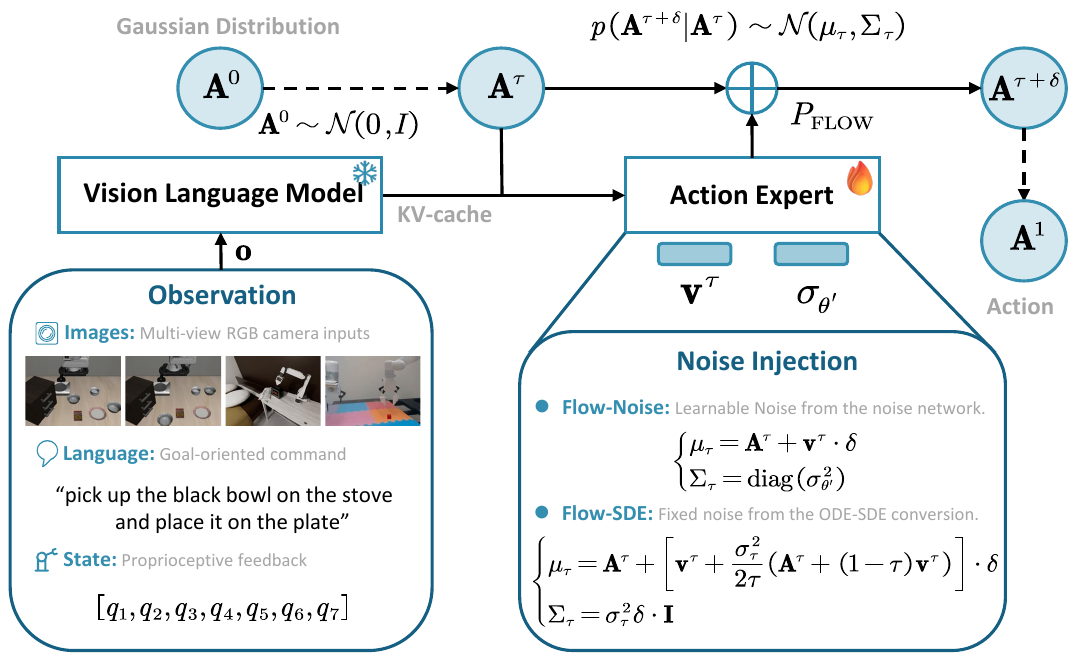}
    \caption{Illustration for the noise injection in $\pi_{\texttt{RL}}$.}
    \label{fig:noise_injection}
\end{figure}

In Flow-Noise, we parameterize the noise schedule with a neural network, allowing the magnitude of the injected noise to be learned dynamically during training for greater flexibility, as shown in \cref{fig:noise_injection}. We focus on the generation process within a single environment timestep $t$. For notational simplicity, we omit the time subscript $t$, \eg, writing $\mathbf{A}^\tau$, and denote the predicted velocity $\mathbf{v}_{\theta}(\mathbf{A}^{\tau}, \mathbf{o})$ as $\mathbf{v}^{\tau}$.

The step transition during the denoising process is modeled as a Gaussian distribution $p(\mathbf{A}^{\tau+\delta} | \mathbf{A}^\tau) \sim \mathcal{N}(\mu_\tau, \Sigma_\tau)$, where the mean is determined by the forward Euler update of the original ODE and the variance is controlled by the learnable noise network $\theta'$:
\begin{equation}
\begin{cases}
\label{equ:policy_transition}
    \mu_\tau = \mathbf{A}^\tau + \mathbf{v}^\tau \cdot \delta \\
    \Sigma_\tau = \text{diag}(\sigma_{\theta'}^2) 
\end{cases}.
\end{equation}
Here, $\sigma_{\theta'}(\cdot)$ is the standard deviation learned from the noise injection network, conditioned on the action $\mathbf{A}^\tau$, and the observation $\mathbf{o}$. The noise network is trained jointly with the velocity but discarded after fine-tuning, leaving a deterministic policy for inference.

\subsubsection{Log-Likelihood Estimation}
\label{subsubsec:one_mdp}
The primary challenge in applying policy gradient methods to flow-based VLAs stems from the intractable log-likelihood of the final executed action. In Flow-Noise, we address it by substituting the gradient of the joint log-likelihood of the entire denoising process into the policy optimization objective in \cref{equ:policy_opt}, which is theoretically grounded in Reinflow \citep{zhang2025reinflow}. 

\begin{figure}
    \centering
    \includegraphics[width=0.95\linewidth]{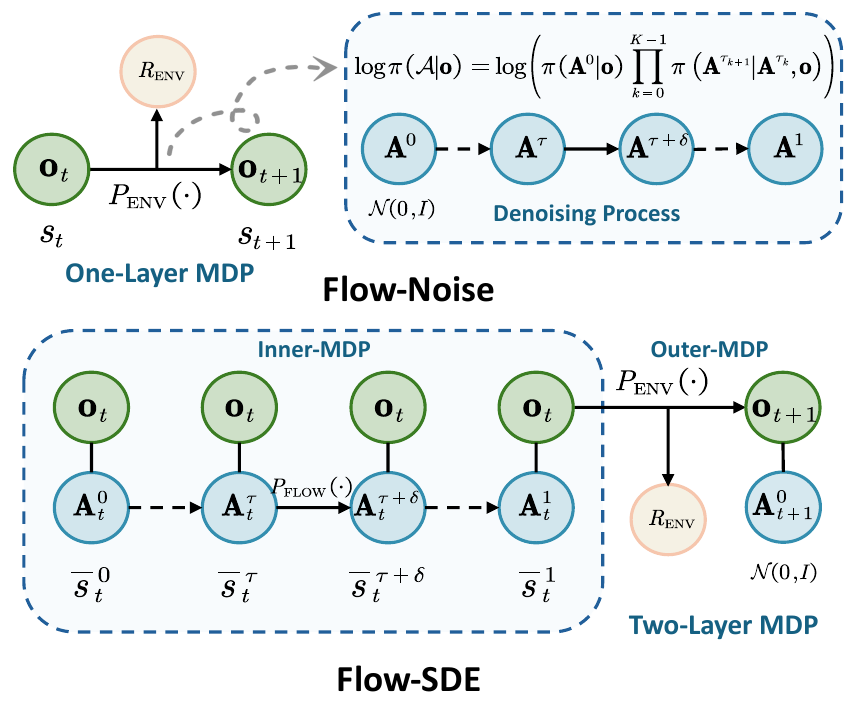}
    \caption{
    Illustration of the MDP formulations in $\pi_{\texttt{RL}}$. 
    }
    \label{fig:framework}
\end{figure}

The inference process for action generation is discretized into $K$ uniform steps, which defines a sequence of time points $\{\tau_0, \tau_1, \dots, \tau_{K}\}$. With the step interval defined as $\delta=1/K$, the discrete timestep at the $k$-th point is $\tau_k = k \cdot \delta$, starting from  $\tau_0 = 0$ and culminating at $\tau_{K} = 1$. Given the observation $\mathbf{o}$, the exact and tractable log probability for the entire denoising sequence $\mathcal{A} = (\mathbf{A}^0,\dots,\mathbf{A}^1)$ is depicted in \cref{fig:framework} and formulated as:
\begin{equation}
\log \pi(\mathcal{A}|\mathbf{o}) = \log \left( \pi(\mathbf{A}^0|\mathbf{o}) \prod_{k=0}^{K-1} \pi(\mathbf{A}^{\tau_{k+1}}| \mathbf{A}^{\tau_k}, \mathbf{o}) \right).
\end{equation}

Building on this, we can treat flow-based policy optimization within a standard MDP framework.

\subsection{Flow-SDE}
Inspired by Flow-GRPO \citep{liu2025flowgrpo}, we enhance stochastic exploration by converting the denoising process from ODE into an SDE formulation. We further construct a two-layer MDP to couple the denoising process with the policy-environment interaction following DPPO \citep{ren2024dppo}, while leveraging the hybrid ODE-SDE sampling technique to accelerate the training process. 

\subsubsection{Stochasticity Injection}
\label{subsubsec:fixed_noise}
In Flow-SDE, we convert the deterministic ODE into an equivalent SDE that preserves the marginal probability density of the generated actions, as shown in \cref{fig:noise_injection}. 

The deterministic ODE sampling trajectory of the flow matching, especially the Rectified Flow \citep{liu2022recflow}, is described by the forward Euler method:
\begin{equation}
    d\mathbf{A}^\tau = \mathbf{v}^{\tau} d\tau.
    \label{equ:flow_matching_ode}
\end{equation}

Building on the connection between the \textit{probability flow} ODE and SDE \citep{song2020score}, we can transform the deterministic ODE in \cref{equ:flow_matching_ode} into an equivalent SDE, with a drift term that corrects the original velocity and a diffusion term that introduces noise:
\begin{equation}
    d\mathbf{A}^\tau = \underbrace{\left( \mathbf{v}^\tau - \frac{1}{2}g^2(\tau) \nabla\log q_{\tau}(\mathbf{A}^{\tau}) \right) d\tau}_{\text{Drift Term}} + \underbrace{g(\tau) d\mathbf{w}}_{\text{Diffusion Term}},
    \label{equ:sde_format}
\end{equation}
where $g(\tau)$ is a scalar function controlling the noise schedule, $\nabla\log q_{\tau}(\mathbf{A}^{\tau})$ is the score function of the marginal distribution $q_{\tau}$ and $d\mathbf{w}$ denotes a Wiener process.

As established in Flow-GRPO, the score function and the velocity field are critically linked by
$
    \nabla \log q_\tau(\mathbf{A}^\tau) = -\frac{\mathbf{A}^\tau}{\tau} - \frac{1-\tau}{\tau} \mathbf{v}^\tau
$.
By substituting the score function with the velocity field in \cref{equ:sde_format} and setting the noise schedule $g(\tau)$ to $\sigma_\tau=a\sqrt{\frac{\tau}{1-\tau}}$ with $a$ controlling the noise level, we derive the final SDE formulation for the flow-matching sampler:
\begin{equation}
    d\mathbf{A}^\tau = \left[ \mathbf{v}^\tau + \frac{\sigma_\tau^2}{2\tau} \left( \mathbf{A}^\tau + (1-\tau) \mathbf{v}^\tau \right) \right] d\tau + \sigma_\tau d\mathbf{w}_\tau.
    \label{equ:sde_final}
\end{equation}

Discretizing this SDE reveals that the transition probability $p(\mathbf{A}^{\tau+\delta} | \mathbf{A}^\tau) \sim \mathcal{N}(\mu_\tau, \Sigma_\tau)$ is an isotropic Gaussian distribution, with the mean and variance formulated as:
\begin{equation}
\begin{cases}
    \mu_\tau = \mathbf{A}^\tau + \left[ \mathbf{v}^\tau + \frac{\sigma_\tau^2}{2\tau} \left( \mathbf{A}^\tau + (1-\tau)\mathbf{v}^\tau \right) \right] \cdot \delta \\
    \Sigma_\tau = \sigma_\tau^2 \delta \cdot \mathbf{I}
\end{cases}.
\end{equation}

\subsubsection{MDP Formulation}
\label{subsubsec:bilevel_mdp}
We couple the denoising process of the flow matching with environmental interaction in Flow-SDE. Specifically, we embed the inner MDP defined during the denoising process into the high-level, outer-loop MDP with the environment $\mathcal{M}_{\text{ENV}}$ in \cref{subsec:mdp}, formulating a two-layer MDP as shown in \cref{fig:framework}, with components defined with respect to the environment time $t$ and denoising time $\tau$.

\begin{itemize}
    \item \textbf{State} $\bar{s}_t^\tau = (\mathbf{o}_t, \mathbf{A}_t^\tau)$ is the tuple of the observation $\mathbf{o}_t$ and the action state $\mathbf{A}_t^\tau$. 

    \item \textbf{Action} $\bar{a}_t^\tau$ is defined as the next sampled denoised action in the inner-loop and the executed action for the outer loop:
    \begin{equation}
        \label{equ:mdp_action}
        \bar{a}_t^\tau = 
        \begin{cases}
            \mathbf{A}_t^{\tau + \delta} & \text{if } \tau < 1  \\
            \mathbf{A}_t^1 & \text{if } \tau = 1 
        \end{cases},
    \end{equation}
    where $\mathbf{A}_t^{\tau + \delta}=\mu_\tau + \sigma_\tau \sqrt{\delta} \cdot \boldsymbol{\epsilon}$, $\boldsymbol{\epsilon} \sim \mathcal{N}(0, \mathbf{I})$ is the randomly sampled noise.
    
    \item \textbf{Transition} $\bar{P}(\bar{s}_{t'}^{\tau'} | \bar{s}_t^\tau, \bar{a}_t^\tau)$ defines how the state evolves, formulated as:
    \begin{equation}
        \bar{s}_{t'}^{\tau'} = 
        \begin{cases} 
            (\mathbf{o}_t, \bar{a}_t^\tau) & \text{if } \tau < 1 \\
            (\mathbf{o}_{t+1}, \mathbf{A}_{t+1}^0) & \text{if } \tau = 1 
        \end{cases}.
    \end{equation}
    For $\tau < 1$, the inner loop transition $P_{\text{FLOW}}(\cdot)$ occurs between different denoised action states, where the observation $\mathbf{o}_t$ remains fixed and the next action state is set by $\bar{a}_t^\tau=\mathbf{A}_t^{\tau+\delta}$. 
    
    For $\tau = 1$, the final action $\bar{a}_t^\tau=\mathbf{A}_t^1$ interacts with the outer-loop environment, resulting in a new observation $\mathbf{o}_{t+1}$ according to the environment dynamics $P_{\text{ENV}}(\cdot)$. Concurrently, the action state is reset from a standard normal distribution $\mathbf{A}_{t+1}^0 \sim \mathcal{N}(0, I)$.
    
    \item \textbf{Reward} $\bar{R}(\bar{s}_t^\tau, \bar{a}_t^\tau)$ is granted only upon completion of the denoising process and interaction with the environment:
    \begin{equation}
        \bar{R}(\bar{s}_t^\tau, \bar{a}_t^\tau) =
        \begin{cases}
            0 & \text{if } \tau < 1 \\
            R_{\text{ENV}}(\mathbf{o}_t, \mathbf{A}_t^1) & \text{if } \tau = 1
        \end{cases}.
    \end{equation}
\end{itemize}

Within the two-layer MDP framework, the problem of estimating the action log-likelihood $\log\pi(a_t|s_t)$ is transformed into estimating $\log\pi(\bar{a}_t^{\tau}|\bar{s}_t^{\tau})$, which is straightforward to compute due to the Gaussian nature of the transitions.

\begin{figure*}[t] 
    \centering
    \begin{subfigure}{0.493\textwidth}
        \centering
        \includegraphics[width=\linewidth]{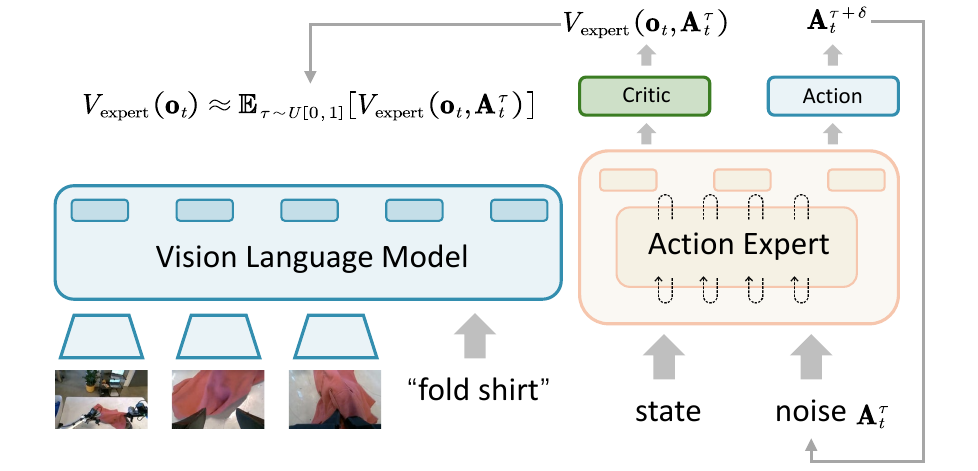}
        \caption{Critic with the action expert, exemplified by $\pi_{0}$.}
    \end{subfigure}%
    \begin{subfigure}{0.493\textwidth}
        \centering
        \includegraphics[width=\linewidth]{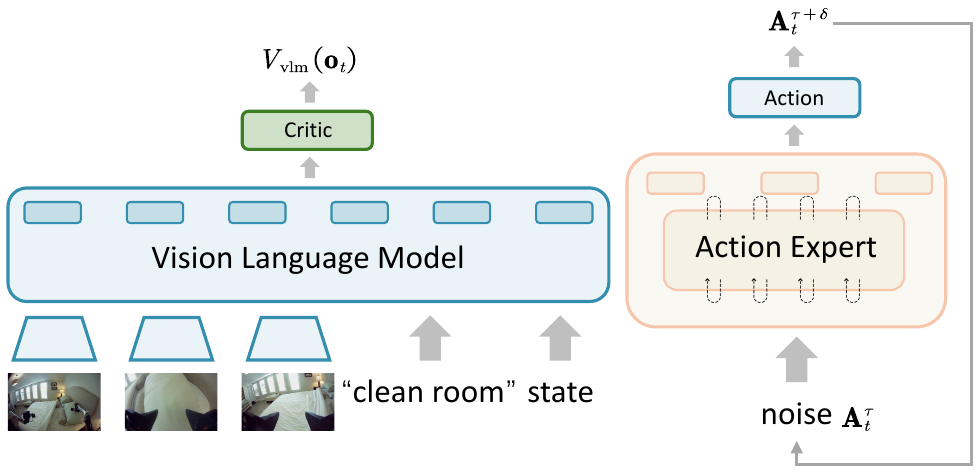}
        \caption{Critic with the VLM, exemplified by $\pi_{0.5}$.}
    \end{subfigure}
    \caption{Illustration of the two critic placement configurations.}
    \label{fig:critic_design}
\end{figure*}

\subsubsection{Hybrid ODE-SDE Sampling}

The two-layer MDP formulation significantly extends the horizon, increasing training difficulty and computational cost. To mitigate this, we adopt a mixed ODE-SDE rollout strategy \citep{li2025mixgrpo, he2025tempflow}. At each step $t$, we randomly sample a denoising time $\tau_t$ for the stochastic SDE exploration, while treating all remaining denoising steps as deterministic ODE updates. Specifically, the policy acts on the state $\bar{s}_t^{\tau_t} = (\mathbf{o}_t, \mathbf{A}_t^{\tau_t})$; subsequently, an environment wrapper executes the remaining ODE steps and the environment transition, ultimately yielding the next state $\bar{s}_{t+1}^{\tau_{t+1}} = (\mathbf{o}_{t+1}, \mathbf{A}_{t+1}^{\tau_{t+1}})$ at a newly sampled time $\tau_{t+1}$. This formulation effectively shortens the MDP horizon while maintaining theoretical consistency with the original two-layer framework.

\subsection{Policy Optimization}
\label{subsec:optimization}
\subsubsection{Algorithm}
Given the formulated flow policy MDP, our objective is to learn the optimal parameters $\theta^*$ for the policy $\pi_\theta$ that maximizes the expected discounted return $\mathcal{J}(\pi_\theta)$. To this end, we apply the widely adopted policy gradient algorithm PPO to optimize the policy.

$\pi$-series models \citep{black2024pi_0,intelligence2025pi05} adopt a chunk-based approach for action generation. Specifically, the policy outputs an entire sequence of $H$ future actions $\mathbf{A}_t = [a_{t,0}, ..., a_{t,H-1}]$ in response to each observation. In this approach, we treat the entire sequence as a single macro-step and define its corresponding reward $R_t = \sum_{j=0}^{H-1} r_{t,j}$ as the sum of the per-step rewards $r_{t,j}$, referred to as the chunk-level formulation in RLinf-VLA \citep{zang2025rlinf}.

To effectively guide policy updates, PPO employs Generalized Advantage Estimation (GAE) \citep{gae} to compute a low-variance estimate of the advantage, estimated as:
\begin{equation}
    \hat{A}_t = \sum_{k=0}^{T-t} (\gamma\lambda)^k \mathcal{T}_{t+k},
\end{equation}
where the TD-error is $\mathcal{T}_t = R_t + \gamma V(s_{t+1}) - V(s_t)$. Here, $V(\cdot)$ is the state-value function derived from the critic network, $\gamma$ is the discount factor, and $\lambda$ is the parameter that balances the trade-off between bias and variance in the advantage estimate.

PPO constrains policy updates to a small trust region to prevent large, destabilizing updates, with the objective function:
\begin{equation}
    \mathcal{J}(\pi_\theta) = \mathbb{E}_t \left[ \min\left( \rho_t(\theta)\hat{A}_t, \ \text{clip}(\rho_t(\theta), 1-\epsilon, 1+\epsilon)\hat{A}_t \right) \right],
\end{equation}
where the $\text{clip}$ function, governed by a hyperparameter $\epsilon$, restricts the ratio $\rho_t(\theta)$ to the interval $[1-\epsilon, 1+\epsilon]$ to ensure training stability.

Here, the probability ratio $\rho_t(\theta)$ between the updated and old policies takes the form of either:
\begin{equation}
    \rho_t(\theta) = \frac{\pi_{\theta_{\text{new}}}(a_t|s_t)}{\pi_{\theta_{\text{old}}}(a_t|s_t)} \quad \text{or} \quad \rho_t(\theta) = \frac{\pi_{\theta_{\text{new}}}(\bar{a}_t^{\tau}|\bar{s}_t^{\tau})}{\pi_{\theta_{\text{old}}}(\bar{a}_t^{\tau}|\bar{s}_t^{\tau})}.
\end{equation}

\subsubsection{Critic Design}
Following VLA-PPO works \citep{zang2025rlinf, rl4vla}, we employ a shared actor-critic architecture for memory-efficient value prediction as shown in \cref{fig:critic_design}. However, the two flow-based VLAs process the proprioceptive state differently: in $\pi_0$, the state is fed into the action expert model, whereas in $\pi_{0.5}$, it is merged with prompt embeddings within the VLM.

To this end, for the $\pi_{0.5}$ variant, we attach the critic network directly to the VLM output, providing the value estimate $V_{\text{vlm}}(\mathbf{o}_t)$ conditioned on the integrated image, language, and state inputs. Conversely, for the $\pi_0$ variant, achieving the value prediction is non-trivial due to the coupled input structure, where the action expert requires both the noisy action $\mathbf{A}_t^\tau$ and the state. 
To this end, we approximate $V_{\text{expert}}(\mathbf{o}_t)$ by averaging the value estimates across the entire denoising trajectory, formulated as:
\begin{equation}
    V_{\text{expert}}(\mathbf{o}_t) \approx \mathbb{E}_{\tau \sim U[0,1]}[V_{\text{expert}}(\mathbf{o}_t, \mathbf{A}_t^\tau)].
    \label{eq:avg_value_est}
\end{equation}

\section{Experimental Results}
\subsection{Setup}
\textbf{Benchmarks.} 
We perform experiments on four widely-adopted robot manipulation benchmarks: \texttt{\textbf{LIBERO}} \citep{liu2023libero}, \texttt{\textbf{ManiSkill}} \citep{tao2024maniskill3}, \texttt{\textbf{MetaWorld}} \citep{mclean2025meta} and \texttt{\textbf{CALVIN}} \cite{mees2022calvin}.
    
\textbf{Flow-based VLAs.}
We conduct our primary experiments based on $\pi_0$ and $\pi_{0.5}$ models. 
Additionally, we conduct experiments on GR00T in Appendix \cref{appendix:gr00t}, which validates that our algorithm can be applied to other flow-based VLAs.

\begin{table*}[t]
\centering
\caption{Comprehensive ID performance comparison across four benchmarks.}
\label{tab:main_results_comparison}
\begin{tabular*}{\textwidth}{@{\extracolsep{\fill}}llcccccl} 
\toprule
\multicolumn{2}{l}{\multirow{2}{*}{\textbf{Model}}} & \multicolumn{6}{c}{\textbf{Benchmarks}} \\ 
\cmidrule(lr){3-8}
\multicolumn{2}{l}{} & \textbf{LIBERO} & \textbf{ManiSkill} & \textbf{MetaWorld} & \textbf{CALVIN} & \textbf{Avg.} & \textbf{$\Delta$ Avg.} \\ 
\midrule  
\hspace{0.38em} \multirow{3}{*}{$\pi_0$} 
    & \hspace{0.38em} SFT         & 57.6 & 38.4 & 50.8 & 57.5 & 51.1 & --- \\ 
    & \hspace{0.38em} Flow-SDE    & 96.1 & \textbf{78.8} & 78.1 & \textbf{61.7} & 78.7 & \deltaimp{27.6} \\ 
    & \hspace{0.38em} Flow-Noise  & \textbf{97.6} & 77.8 & \textbf{85.8} & 59.9 & \textbf{80.3} & \deltaimp{\textbf{29.2}} \\ 
\midrule
\hspace{0.38em} \multirow{3}{*}{$\pi_{0.5}$} 
    & \hspace{0.38em} SFT         & 77.1 & 40.1 & 43.8 & 61.3 & 55.6 & --- \\ 
    & \hspace{0.38em} Flow-SDE    & 97.9 & \textbf{90.9} & \textbf{70.7} & \textbf{87.0} & \textbf{86.6} & \deltaimp{\textbf{31.0}} \\ 
    & \hspace{0.38em} Flow-Noise  & \textbf{98.3} & 89.7 & 66.1 & 84.5 & 84.7 & \deltaimp{29.1} \\ 
\bottomrule
\end{tabular*}
\end{table*}

\begin{figure*}[t] 
    \centering
    \begin{subfigure}{0.25\textwidth}
        \centering
        \includegraphics[width=\linewidth]{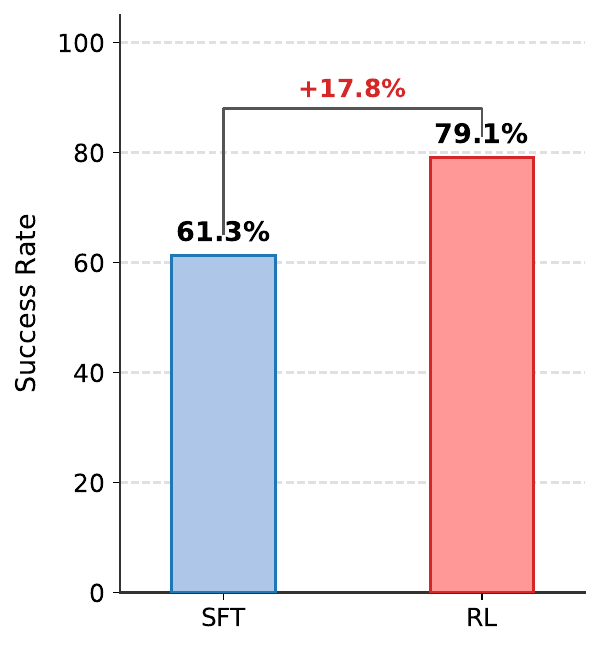}
        \caption{CALVIN}
    \end{subfigure}
    \hspace{2.3em}
    \begin{subfigure}{0.3\textwidth}
        \centering
        \includegraphics[width=\linewidth]{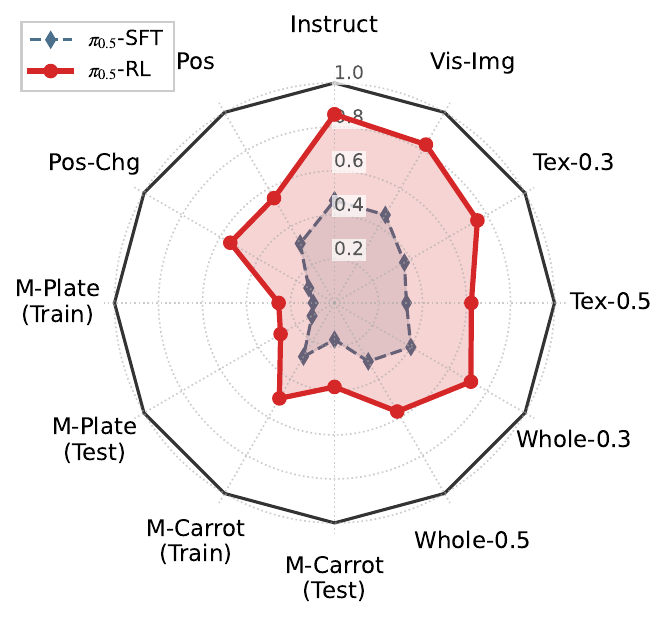}
        \caption{ManiSkill}
    \end{subfigure}
    \hspace{2.3em}
    \begin{subfigure}{0.25\textwidth}
        \centering
        \includegraphics[width=\linewidth]{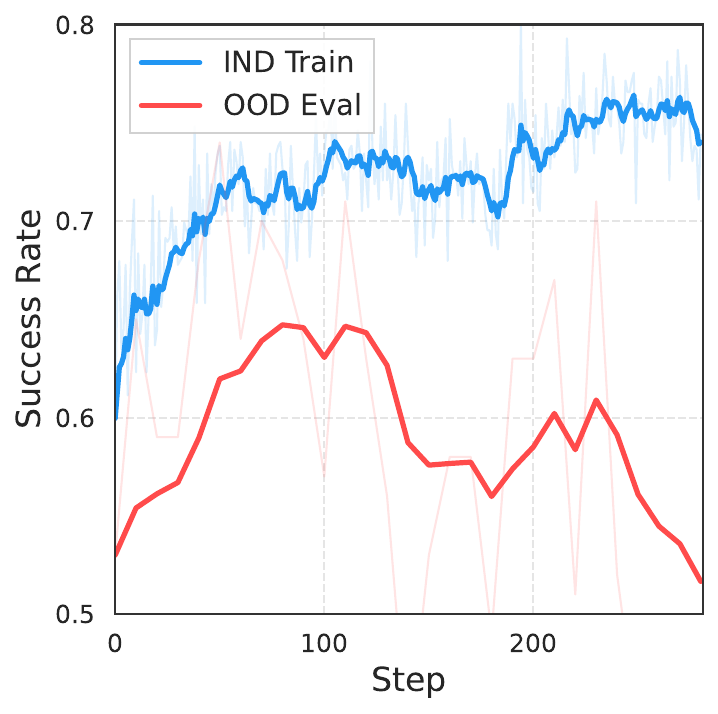}
        \caption{MetaWorld}    
    \end{subfigure}
    
    \caption{Comprehensive OOD evaluation results on CALVIN ABC-D, ManiSkill OOD, and MetaWorld ML45 benchmarks.}
    \label{fig:ood_overall}
\end{figure*}

\subsection{Main Results}
In this section, we assess the in-distribution (ID) performance of $\pi_{\texttt{RL}}$ across various benchmarks, followed by an analysis of its out-of-distribution (OOD) generalization.

\subsubsection{In-Distribution RL Training}
As detailed in \cref{tab:main_results_comparison}, $\pi_{\texttt{RL}}$ yields substantial performance gains over SFT baselines across all evaluated benchmarks. Specifically, the $\pi_0$ model achieves a maximum average improvement of +29.2\%, while the $\pi_{0.5}$ variant demonstrates a +31.0\% increase in average success rate. 

Specifically for LIBERO, we perform few-shot SFT on the $\pi_{0.5}$ model followed by RL optimization to achieve a 98.3\% success rate, outperforming the 96.9\% success rate of the full-dataset SFT baseline. These performance gains extend to other challenging environments, including ManiSkill with its 4,352 pick-and-place task combinations, MetaWorld featuring 50 distinct manipulation primitives, and CALVIN for long-horizon sequential tasks. See Appendix \cref{appendix:ind_experiments} for comprehensive experimental details.

\subsubsection{Out-of-Distribution RL Evaluation}
While previous experiments demonstrate that RL yields performance improvements, a critical question remains: \textit{does RL yield an enhanced policy, or simply overfit to the ID environment driven by provided rewards?} In this section, we evaluate RL-finetuned policies in OOD scenarios, where the environment distribution or the task objective deviates from the ID training setup, to assess their generalization capabilities.

As illustrated in \cref{fig:ood_overall}, the performance gains achieved in the ID setting effectively transfer to OOD scenarios in ManiSkill and CALVIN, where the domain shift primarily stems from environmental variations. Conversely, for the OOD setting in MetaWorld, which involves distinct manipulation tasks, performance fluctuates without showing significant improvement. This finding suggests that the benefits of RL are primarily localized to action-level refinement rather than broader augmentation of cross-task generalization capabilities. See Appendix \cref{appendix:ood_experiments} for more details.

\subsection{Ablation Study}
Given that Flow-SDE achieves performance comparable to Flow-Noise while offering higher computational efficiency, we conduct our ablation studies with the Flow-SDE method. Specifically, we investigate the impact of critic designs, noise injection strategies, and MDP formulations, with additional results on RL algorithms and hyper-parameters provided in Appendix \cref{appendix:ablation}.

\begin{figure*}[!t] 
    \centering
    \begin{subfigure}{0.32\textwidth}
        \centering
        \includegraphics[width=\linewidth]{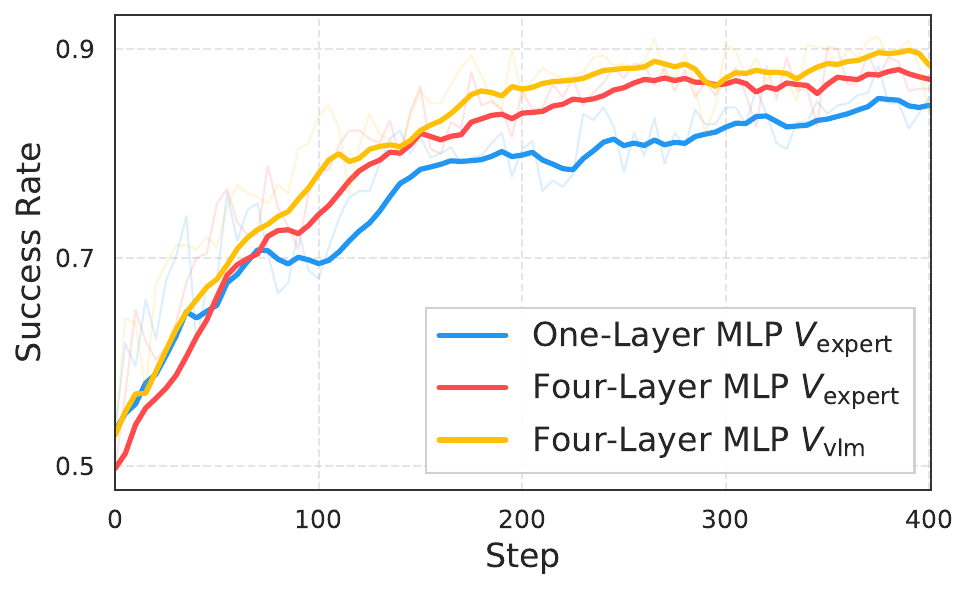}
        \caption{Eval}
    \end{subfigure}%
    \begin{subfigure}{0.32\textwidth}
        \centering
        \includegraphics[width=\linewidth]{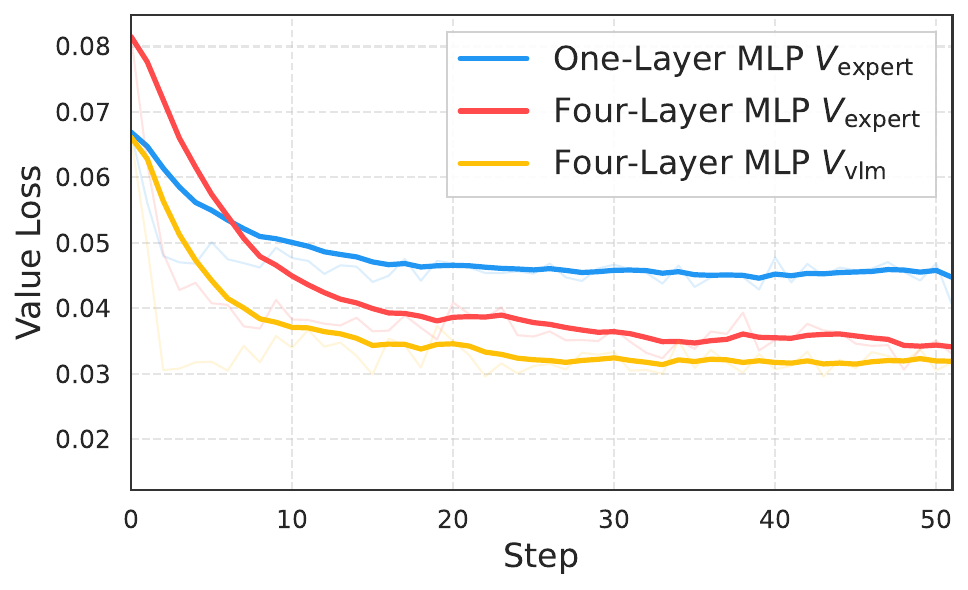}
        \caption{Value Loss}
    \end{subfigure}%
    \begin{subfigure}{0.32\textwidth}
        \centering
        \includegraphics[width=\linewidth]{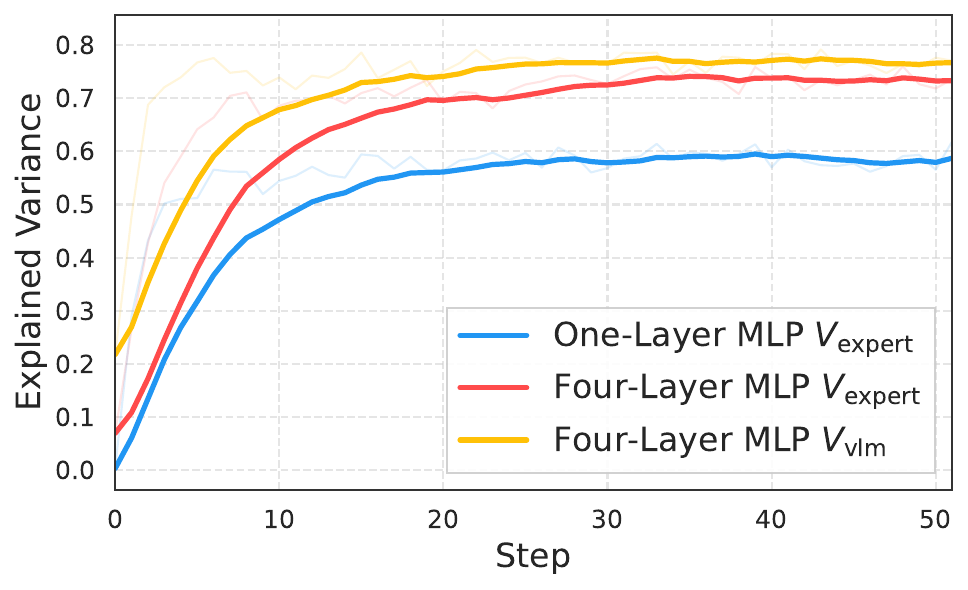}
        \caption{Explained Variance}
    \end{subfigure}
    \caption{
    Ablation on the critic design within Flow-SDE $\pi_0$ on the LIBERO-Long, indicating that the critic $V_{\text{vlm}}$ attached after the VLM exhibits superior performance. Furthermore, a four-layer MLP demonstrates stronger regression capability in $V_{\text{expert}}$.
    }
    \label{fig:ablation_critic_design}
\end{figure*}
\begin{figure*}[!t] 
    \centering
    \begin{subfigure}{0.32\textwidth}
        \centering
        \includegraphics[width=\linewidth]{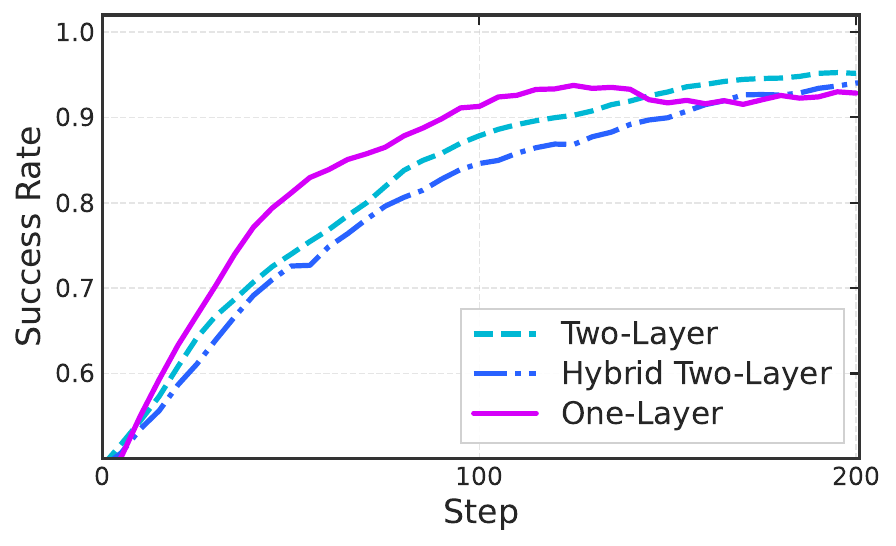}
        \caption{Eval}
    \end{subfigure}%
    \begin{subfigure}{0.32\textwidth}
        \centering
        \includegraphics[width=\linewidth]{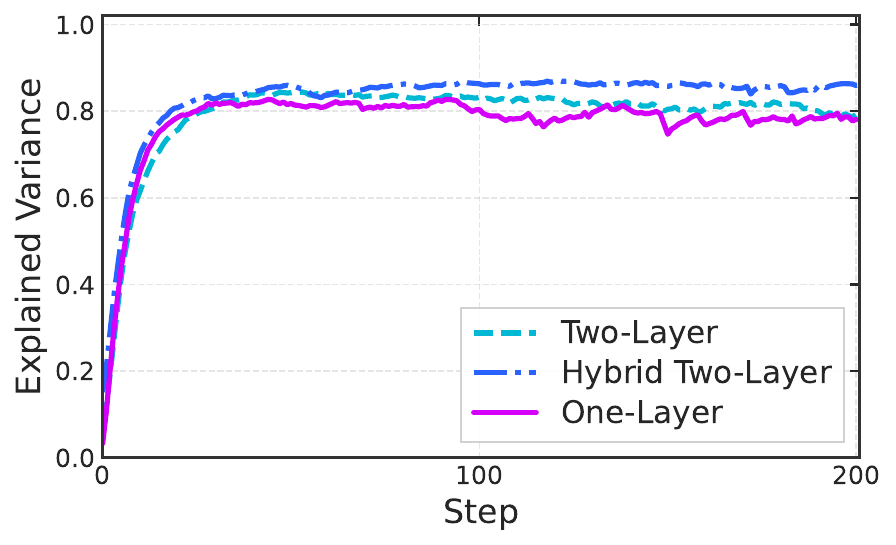}
        \caption{Explained Variance}
    \end{subfigure}
    \begin{subfigure}{0.32\textwidth}
        \centering
        \includegraphics[width=\linewidth]{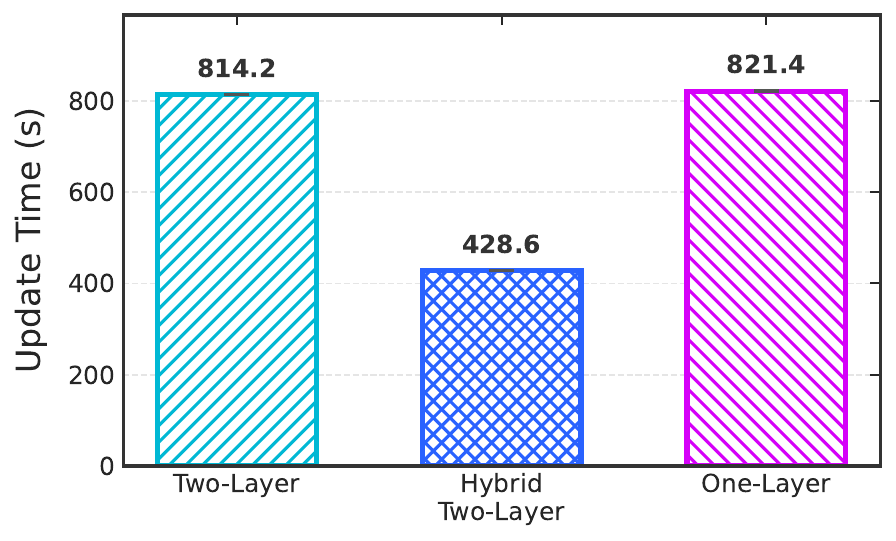}
        \caption{Update Time}
    \end{subfigure}
    \caption{Ablation on the MDP formulation within Flow-SDE of $\pi_0$ on the LIBERO-Goal.}
    \label{fig:ablation_mdp}
\end{figure*}
\begin{figure*}[!t] 
    \centering
    \begin{subfigure}{0.32\textwidth}
        \centering
        \includegraphics[width=\linewidth]{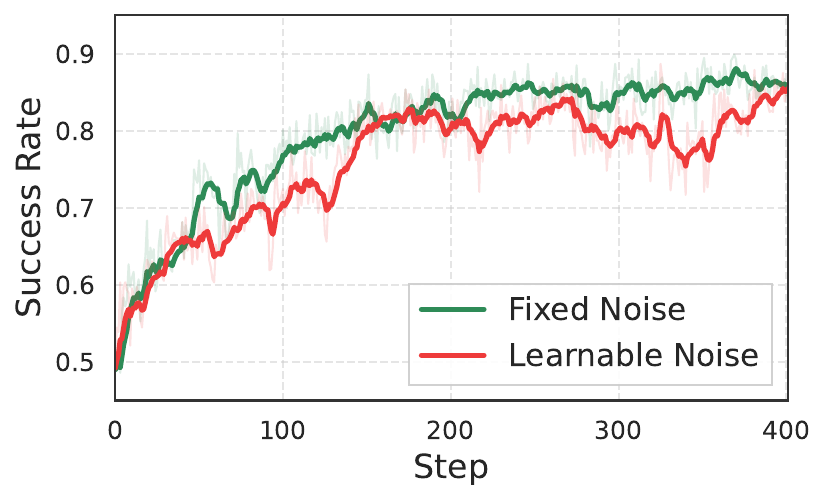}
        \caption{Train}
    \end{subfigure}%
    \hspace{5em}
    \begin{subfigure}{0.32\textwidth}
        \centering
        \includegraphics[width=\linewidth]{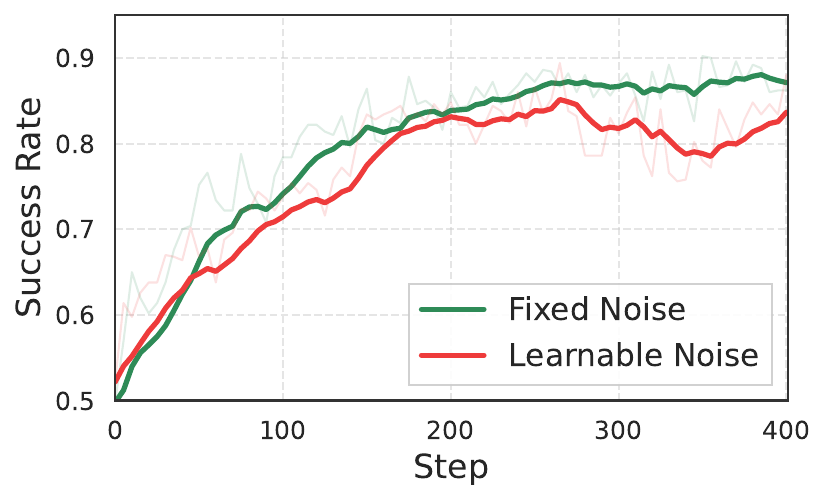}
        \caption{Eval}
    \end{subfigure}%
    \caption{Ablation on the injection strategy within Flow-SDE of $\pi_0$ on the LIBERO-Long.}
    \label{fig:ablation_noise_injection}
\end{figure*}

\subsubsection{Critic Design}
\textbf{Placement.}
We compare two critic placement strategies, one positioned after the action expert ($V_{\text{expert}}$) and the other after the VLM ($V_{\text{vlm}}$), with $\pi_0$ model on the LIBERO-Long task suite. As illustrated in \cref{fig:ablation_critic_design}, we observe that $V_{\text{vlm}}$ exhibits slightly superior performance, lower value loss, and higher explained variance, despite not receiving the proprioceptive state as input. This advantage can be attributed to a key difference in their input: $V_{\text{vlm}}$ learns a direct mapping from observation to value, while $V_{\text{expert}}$ must contend with optimization challenges arising from coupled state and noisy action inputs. 

Nevertheless, to align with the concept of the value function, we maintain the $V_{\text{expert}}$ architecture for the $\pi_0$, ensuring that state information is incorporated to estimate the value.

\textbf{Structure.}
We investigate a four-layer MLP versus a one-layer MLP, which mirrors the action-projection structure in the action expert. Results in \cref{fig:ablation_critic_design} indicate that the four-layer MLP leads to a more accurate value approximation, resulting in enhanced performance and training stability.

\subsubsection{Flow Policy MDP}
With the same fixed noise injection strategy, we evaluate the one-layer MDP of Flow-Noise with the two-layer MDP of Flow-SDE on the LIBERO-Goal, as shown in \cref{fig:ablation_mdp}.

\begin{table*}[tbp]
\caption{Ablation study of hyperparameters for Flow-SDE on the LIBERO-Spatial. \textit{Train} refers to policy performance during the stochastic rollout phase, whereas \textit{Eval} refers to performance during the deterministic evaluation phase.}
\label{tab:ablation_hyperparams}
\centering
\begin{tabular*}{\dimexpr 1.0\textwidth}{@{\extracolsep{\fill}}cclcccccccccc@{}}
\toprule
\multirow{3}{*}[-1.5ex]{\textbf{Models}} & \multirow{3}{*}[-1.5ex]{\textbf{Stage}} & \multicolumn{10}{c}{\textbf{Hyperparameters}} \\
\cmidrule(lr){3-12}
& & \multicolumn{3}{c}{\textbf{Noise Level}} & \multicolumn{4}{c}{\textbf{Denoise Step}} & \multicolumn{3}{c}{\textbf{Action Chunk}} \\
\cmidrule(lr){3-5} \cmidrule(lr){6-9} \cmidrule(lr){10-12}
& & 0.2 & 0.5 & 0.8 & 1 & 2 & 4 & 8 & 5 & 10 & 20 \\
\midrule
\multirow{2}{*}{SFT} 
& Train & 62.3 & 56.0 & 46.6 & 9.4 & 28.3 & 56.1 & 62.6 & 56.0 & 60.7 & 70.3 \\
& Eval & 65.2 & 65.2 & 65.2 & 63.8 & 64.9 & 65.2 & 63.2 & 65.2 & 70.5 & 72.6 \\
\midrule
\multirow{2}{*}{RL} 
& Train & 59.5 & 93.5 & 95.3 & 73.8 & 90.8 & 93.5 & 84.3 & 93.5 & 93.3 & 87.5 \\
& Eval & 73.1 & 94.5 & 98.1 & 88.5 & 97.0 & 94.5 & 86.7 & 94.5 & 95.5 & 89.2 \\
\bottomrule
\end{tabular*}
\end{table*}

\begin{figure*}[t!] 
    \centering
    \begin{subfigure}{0.32\textwidth}
        \centering
        \includegraphics[width=\linewidth]{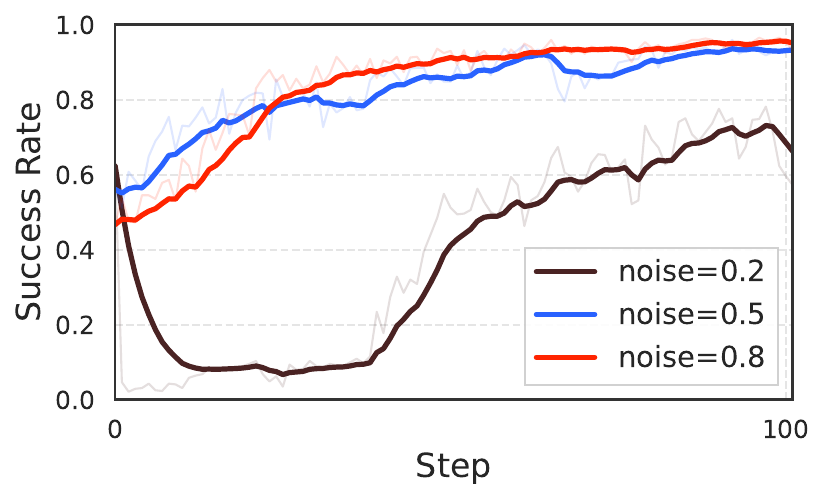}
        \caption{Train}
    \end{subfigure}%
    \hfill
    \begin{subfigure}{0.32\textwidth}
        \centering
        \includegraphics[width=\linewidth]{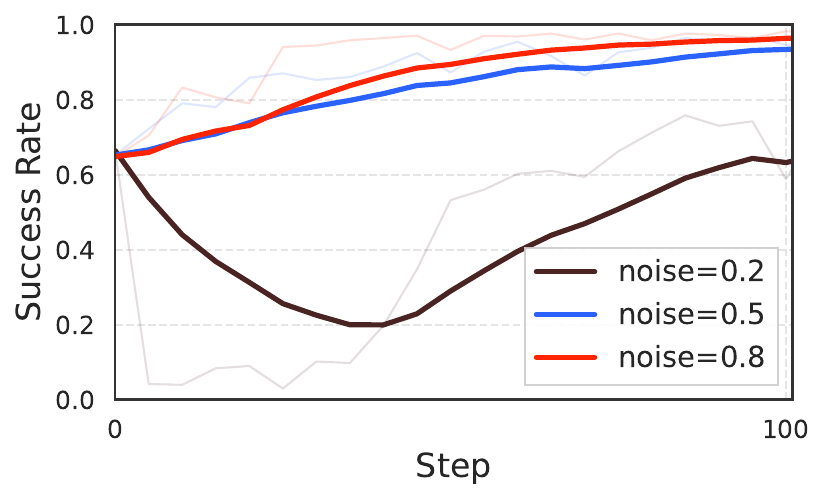}
        \caption{Eval}
    \end{subfigure}%
    \hfill%
    \begin{subfigure}{0.32\textwidth}
        \centering
        \includegraphics[width=\linewidth]{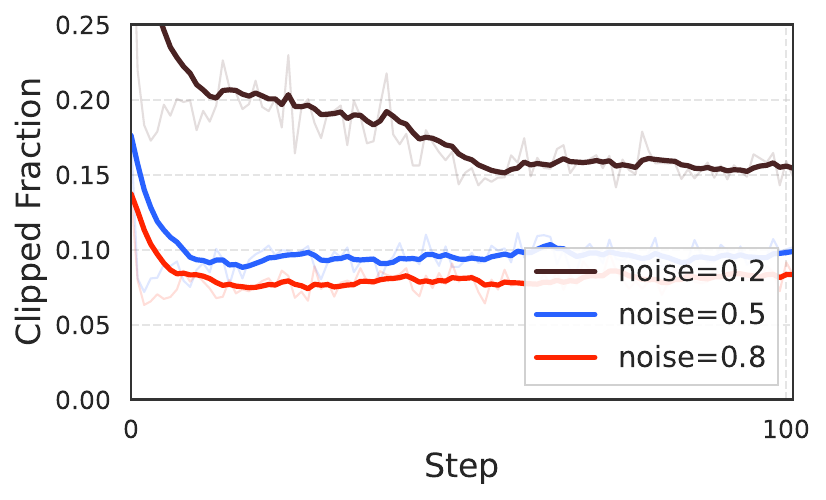}
        \caption{Clipped Fraction}
    \end{subfigure}
    \caption{Ablation on the noise level $a$, conducted with the Flow-SDE $\pi_0$ on the LIBERO-Spatial.}
    \label{fig:ablation_noise}
\end{figure*}

\begin{figure*}[t!] 
    \centering
    \begin{subfigure}{0.32\textwidth}
        \centering
        \includegraphics[width=\linewidth]{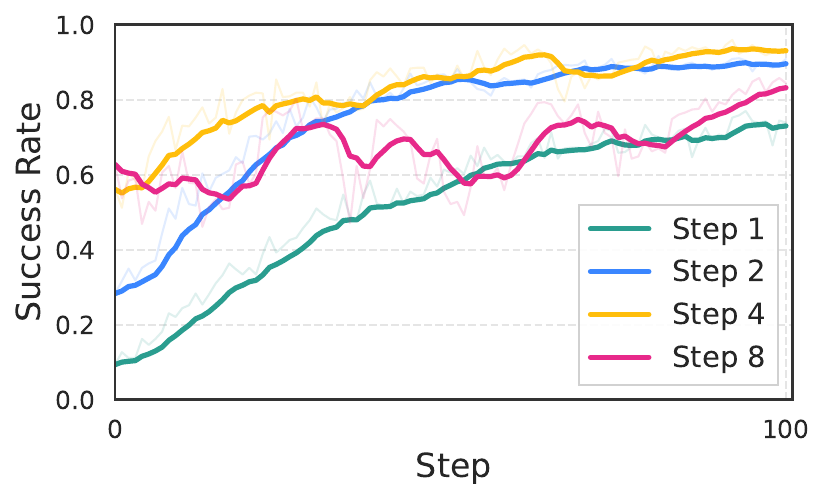}
        \caption{Train}
    \end{subfigure}%
    \hspace{5em} 
    \begin{subfigure}{0.32\textwidth}
        \centering
        \includegraphics[width=\linewidth]{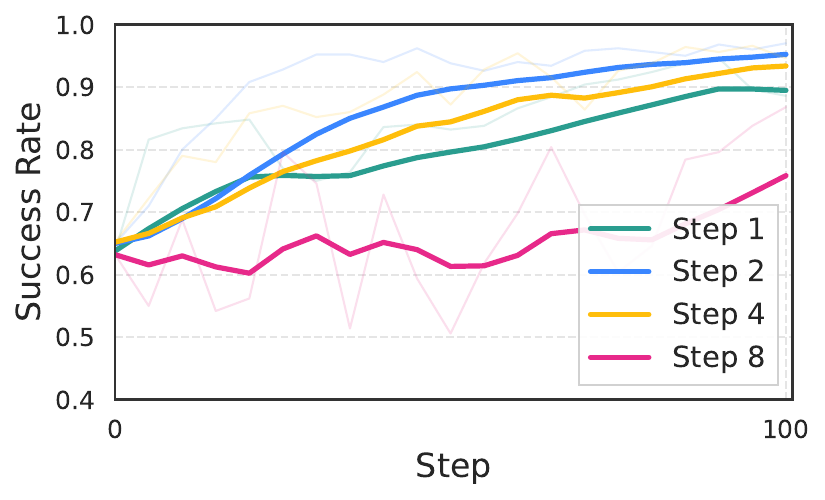}
        \caption{Eval}
    \end{subfigure}%
    \caption{Ablation on the denoise step, conducted with the Flow-SDE $\pi_0$ on the LIBERO-Spatial.}
    \label{fig:ablation_denoisestep}
\end{figure*}

\begin{figure*}[t!] 
    \centering
    \begin{subfigure}{0.32\textwidth}
        \centering
        \includegraphics[width=\linewidth]{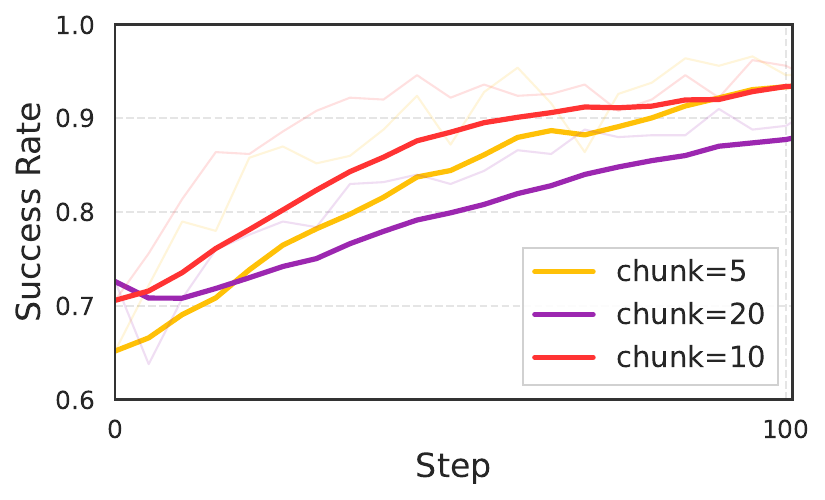}
        \caption{Eval}
    \end{subfigure}%
    \hspace{5em} 
    \begin{subfigure}{0.32\textwidth}
        \centering
        \includegraphics[width=\linewidth]{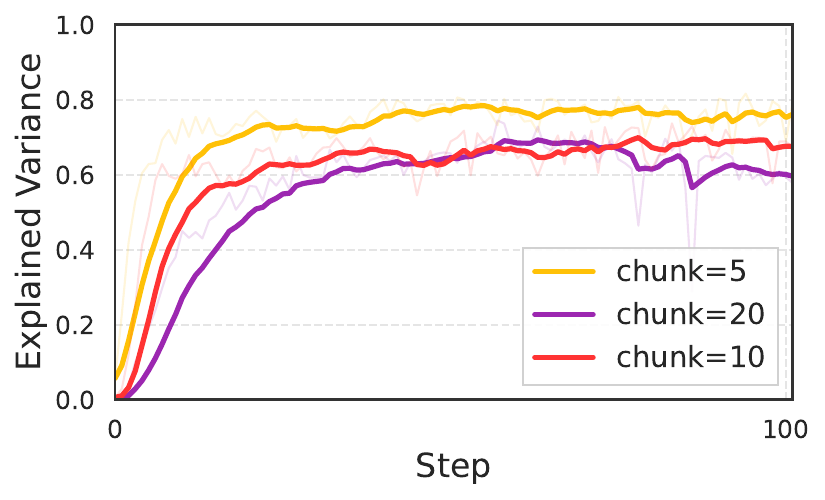}
        \caption{Explained Variance}
    \end{subfigure}%
    \caption{Ablation on the chunk size, conducted with the Flow-SDE $\pi_0$ on the LIBERO-Spatial.}
    \label{fig:ablation_chunk}
\end{figure*}

We observe that the one-layer formulation converges most rapidly, but the final success rates remain consistent across all three formulations. In terms of computational efficiency, the hybrid two-layer paradigm achieves a $2\times$ speedup over the standard approach due to its shorter effective MDP chain. Notably, the one-layer MDP yields no substantial wall-clock time advantage over the standard two-layer model, stemming from the requirement to recalculate full denoising trajectories for log likelihood estimation.

\subsubsection{Stochasticity Injection}
We compare fixed and learnable noise injection strategies using the Flow-SDE MDP formulation on the LIBERO-Long suite. To ensure a controlled comparison, we set the entropy bonus for the learnable noise to zero, aligning it with the fixed noise approach.

Specifically, we set the fixed noise to 0.5, and lower and upper bounds for the learnable noise log-variance to 0.08 and 0.16. As depicted in \cref{fig:ablation_noise_injection}, two noise strategies exhibit similar train performance at step 0, which indicates comparable noise magnitudes. Furthermore, the converged performance affirms the efficiency of both injection methods.

\subsection{Hyper-Parameters}
\label{subsec:hyperparameters}
Building on the Flow-SDE with $\pi_0$ model, we investigate the influence of the noise level, denoise step, and action chunk on the LIBERO-Spatial benchmark. We denote the train stage as the phase where the policy generates stochastic actions for exploration, whereas the evaluation stage involves generating deterministic actions. The train and eval success rates for the SFT baseline and the RL fine-tuned model after 100 training steps are presented in \cref{tab:ablation_hyperparams}.

\textbf{Noise Level.} 
The noise level $a$ in the Flow-SDE is defined in \cref{equ:sde_final}, which governs the noise injection magnitude during the denoising process. As shown in \cref{tab:ablation_hyperparams}, the evaluation performance of the SFT baseline remains identical across all noise levels due to its reliance on deterministic ODE sampling. Conversely, its training performance exhibits a clear degradation as noise increases. This is intuitive, as higher noise levels can distort the flow trajectory, leading to an inaccurate estimation of the marginal action distribution.

Extending this analysis to the RL fine-tuning stage highlights a critical trade-off: \textit{while lower noise levels mitigate exploration-induced performance degradation, they simultaneously constrain the capacity for RL refinement.} This trade-off is empirically supported by \cref{fig:ablation_noise}, which shows that training with minimal noise ($a = 0.2$) leads to instability, characterized by a significantly higher clip fraction. We attribute this instability to the substantially larger gradient magnitudes associated with low-noise regimes.

\textbf{Denoise Step.}
The denoise step $K$ defines the number of discretization steps for action generation and is critical for controlling the fidelity of the ODE-to-SDE transition in \cref{equ:sde_final}. In \cref{tab:ablation_hyperparams}, we observe that while all configurations start with similar eval performance, the train success rate plummets at $K=1$, indicating a significant ODE-to-SDE discretization error.

However, a larger $K$ is not unequivocally optimal. As illustrated in \cref{fig:ablation_denoisestep}, increasing $K$ presents a distinct trade-off: \textit{while it enhances rollout performance, it simultaneously increases training complexity and computational overhead due to the extended sequence of denoising steps.}

\textbf{Action chunk.} 
The action chunk refers to the number of consecutive actions the policy executes within a single observation. We ablate the action chunk size across 5, 10, and 20, with results visualized in \cref{fig:ablation_chunk}. 

Although a larger chunk size yields marginal performance gains, it inherently reduces the frequency of policy-environment interactions and obscures precise reward credit assignment. These constraints lead to less reliable advantage estimation, as evidenced by the diminished explained variance. Consequently, while an increased chunk size may offer a superior SFT baseline, it paradoxically limits the ceiling for subsequent RL-driven refinement.

\section{Conclusion}
We introduce $\pi_{\texttt{RL}}$, a framework that enables flow-based VLAs, $\pi_0$ and $\pi_{0.5}$, to be fine-tuned with online RL algorithms. We tackle the fundamental challenge of intractable log-likelihoods in flow matching with Flow-Noise and Flow-SDE solutions. Our extensive experiments on the challenging benchmarks demonstrated that $\pi_{\texttt{RL}}$ achieves significant performance improvements over SFT baselines.

\textbf{Limitation.} Due to the low sample efficiency of online RL, our framework currently relies on sim-to-real deployment. We aim to develop more efficient algorithms to enable real-world RL training in the future.

\bibliography{example_paper}
\bibliographystyle{icml2026}

\clearpage
\appendix

\section{Appendix}
This appendix provides additional technical details and experimental results for $\pi_{\texttt{RL}}$. The content is organized as:

\begin{itemize}
    \item \textbf{Appendix \ref{appendix:experiment_setup}}: Experimental Setup.
    \item \textbf{Appendix \ref{appendix:ind_experiments}}: In-Distribution RL Training.
    \item \textbf{Appendix \ref{appendix:ood_experiments}}: Out-of-Distribution RL Evaluation.
    \item \textbf{Appendix \ref{appendix:case_study}}: Case Studies: Single-Task RL Training.
    \item \textbf{Appendix \ref{appendix:ablation}}: Ablation Details.
    \item \textbf{Appendix \ref{appendix:large_scale_rl_trianing}}: Insights from Large-Scale RL Training.
    \item \textbf{Appendix \ref{appendix:gr00t}}: RL for GR00T N1.5.
    \item \textbf{Appendix \ref{appendix:limiation}}: Limitations and Future Work.
    \item \textbf{Appendix \ref{appendix:hyper_params}}: Training Hyperparameters.
\end{itemize}

\section{Experimental Setup}
\label{appendix:experiment_setup}
\subsection{Benchmarks.}
To rigorously assess the performance and generalization of our framework, we conduct evaluations across four diverse benchmarks, with diverse emphases on robotic capability:

\begin{itemize}
    \item \texttt{\textbf{LIBERO}} focuses on compositional variations and knowledge transfer in tasks. By evaluating across four task suites: \textit{Spatial}, \textit{Object}, \textit{Goal}, and \textit{Long}, it probes the model's ability to transfer base skills to variations in object arrangements and long-horizon tasks.
    \item \texttt{\textbf{ManiSkill}} emphasizes perceptual and execution robustness under massive environmental diversity. We follow the setup of RL4VLA \citep{rl4vla} and built a benchmark with 4,352 unique pick-and-place combinations, which challenges the policy to maintain physical interactions across a vast distribution of objects and receptacles.
    \item \texttt{\textbf{CALVIN}} evaluates long-horizon sequential reasoning and vision-language grounding. It evaluates the model's capacity to execute chains of five random subtasks in a persistent environment, requiring accurate alignment with linguistic instructions.
    \item \texttt{\textbf{MetaWorld}} measures skill breadth and multi-task versatility. It requires a single policy to master 50 semantically distinct manipulation primitives, ranging from simple reaching to complex tool usage.
\end{itemize}

\subsection{Implementation Details.}
Given that pre-trained models often struggle to generalize to task-specific benchmarks, we initiate our process with SFT on expert demonstrations. For the SFT stage, we fine-tune the entire 3.3B model following the official setting. In the subsequent RL stage, we freeze the VLM parameters and exclusively fine-tune the 300M action expert model, driven by GPU memory efficiency and the findings from RL4VLA that RL contributes more significantly to action generalization. We build the whole framework upon the RLinf \citep{yu2025rlinf} codebase, where we adopt a shared, co-located GPU allocation strategy that places the environment, rollout model, and actor model on the same GPU and executes them serially.

For the model configurations, we adhere to the official setting provided by openpi \citep{black2024pi_0,intelligence2025pi05}. In these settings, $\pi_0$ utilizes image, language, and proprioceptive states as input, whereas $\pi_{0.5}$ notably omits state information\footnote{https://github.com/Physical-Intelligence/openpi/issues/687}. 
Our experiments are conducted on 8 NVIDIA H100 80GB GPUs, and detailed training hyperparameters are available in the Appendix \cref{appendix:hyper_params}.

\begin{table*}[t]
    \centering
    \caption{Evaluation results on the LIBERO benchmark, evaluated based on the success rate (\%).}
    \label{tab:comparison_libero}
    \begin{tabular*}{\textwidth}{@{\extracolsep{\fill}}llcccccl@{}} 
        \toprule
        \multicolumn{2}{l}{\multirow{2}{*}{\textbf{Model}}} & \multicolumn{6}{c}{\textbf{LIBERO}} \\ 
        \cmidrule(lr){3-8}
        \multicolumn{2}{c}{}& \textbf{Spatial} & \textbf{Object} & \textbf{Goal} & \textbf{Long} & \textbf{Avg.} & \textbf{$\Delta$ Avg.} \\ 
        \midrule
        \rowcolor{gray!20}
        \multicolumn{8}{l}{\textit{\# Full Dataset SFT}} \\
        \multicolumn{2}{l}{Octo} & 78.9 & 85.7 & 84.6 & 51.1 & 75.1 & --- \\
        \multicolumn{2}{l}{OpenVLA} & 84.7 & 88.4 & 79.2 & 53.7 & 76.5 & --- \\
        \multicolumn{2}{l}{$\pi_{\text{fast}}$} & 96.4 & 96.8 & 88.6 & 60.2 & 85.5 & --- \\
        \multicolumn{2}{l}{OpenVLA-OFT} & 91.6 & 95.3 & 90.6 & 86.5 & 91.0 & --- \\
        \multicolumn{2}{l}{$\pi_0$} & 96.8 & 98.8 & 95.8 & 85.2 & 94.2 & --- \\
        \multicolumn{2}{l}{$\pi_{0.5}$} & 98.8 & 98.2 & 98.0 & 92.4 & 96.9 & --- \\
        \midrule
        \rowcolor{myblue}
        \multicolumn{8}{l}{\textit{\# Few-shot SFT + RL}} \\
        \hspace{0.38em} \multirow{3}{*}{$\pi_0$} &
           \hspace{0.38em} SFT & 65.3 & 64.4 & 49.8 & 51.2 & 57.6 & --- \\
        & \hspace{0.38em} Flow-SDE & 98.4 & 99.4 & 96.2 & 90.2 & 96.1 & \deltaimp{38.5} \\
        & \hspace{0.38em} Flow-Noise & 99.0  & 99.2 & 98.2 & 93.8 & 97.6 & \deltaimp{\textbf{40.0}} \\
        \midrule
        \rowcolor{mygreen}
        \multicolumn{8}{l}{\textit{\# Few-shot SFT + RL}} \\
        \hspace{0.38em} \multirow{3}{*}{$\pi_{0.5}$} &
           \hspace{0.38em} SFT & 84.6 & 95.4 & 84.6 & 43.9 & 77.1 & --- \\
        & \hspace{0.38em} Flow-SDE & 99.6 & 100 & 98.8 & 93.0 & 97.9 & \deltaimp{20.8} \\
        & \hspace{0.38em} Flow-Noise & \textbf{99.6} & \textbf{100} & \textbf{99.6} & \textbf{94.0} & \textbf{98.3} & \deltaimp{21.2} \\
        \bottomrule
    \end{tabular*}
\end{table*}
\begin{table*}[t]
    \centering
    \caption{Evaluation results on the ManiSkill benchmark, with more specific OOD results depicted in \cref{tab:maniskill_ood_results}.}
    \label{tab:ind_ood_comparison}
    \begin{tabular*}{\textwidth}{@{\extracolsep{\fill}}llccccccc@{}} 
        \toprule
        \multicolumn{2}{l}{\multirow{2}{*}{\textbf{Model}}} & \multicolumn{2}{c}{\textbf{IND}} & \multicolumn{5}{c}{\textbf{OOD}} \\ 
        \cmidrule(lr){3-4} \cmidrule(lr){5-9}
        \multicolumn{2}{c}{}& \textbf{Avg.} & \textbf{$\Delta$} & \textbf{Vision} & \textbf{Semantic} & \textbf{Execution} & \textbf{Avg.} & \textbf{$\Delta$} \\ 
        \midrule
        \hspace{0.38em} \multirow{3}{*}{$\pi_0$} &
            \hspace{0.38em} SFT & 38.4 & --- & 32.6 & 8.4 & 13.2 & 18.1 & --- \\
        & \hspace{0.38em} Flow-SDE & 78.8 & \deltaimp{40.4} & 61.1 & 25.4 & 31.5 & 39.3 & \deltaimp{21.2} \\
        & \hspace{0.38em} Flow-Noise & 77.8 & \deltaimp{39.4} & 63.4 & 23.1 & 24.2 & 36.9 & \deltaimp{18.8} \\
        \midrule
        \hspace{0.38em} \multirow{3}{*}{$\pi_{0.5}$} &
            \hspace{0.38em} SFT & 40.1 & --- & 40.2 & 16.6 & 22.4 & 26.4 & --- \\
        & \hspace{0.38em} Flow-SDE & \textbf{90.9} & \deltaimp{\textbf{50.8}} & 68.0 & 34.5 & 45.4 & 49.3 & \deltaimp{22.9} \\
        & \hspace{0.38em} Flow-Noise & 89.7 & \deltaimp{49.6} & \textbf{69.9} & \textbf{35.5} & \textbf{54.9} & \textbf{53.4} & \deltaimp{\textbf{27.0}} \\
        \bottomrule
    \end{tabular*}
\end{table*}

\section{In-Distribution RL Training}
\label{appendix:ind_experiments}
In this section, we evaluate the effectiveness of our framework across various benchmarks to demonstrate its robustness and superior performance in ID settings.

\subsection{LIBERO Benchmark}
\textbf{SFT Procedure.}
The LIBERO benchmark comprises four task suites, each consisting of 10 distinct subtasks. To facilitate few-shot SFT on LIBERO, a minimum of 40 expert demonstration trajectories is necessary to ensure a positive success rate for each subtask across four task suites, thereby guaranteeing a positive optimization signal for the subsequent RL phase. 

We perform few-shot SFT following the official training configurations. For the $\pi_0$ model, we fine-tune on a subset of 58 trajectories sampled from the 1,692 total demonstrations\footnote{https://huggingface.co/datasets/physical-intelligence/libero}, which serves as the initial checkpoint for subsequent RL training on the \textit{Spatial}, \textit{Object}, and \textit{Goal} task suites. For the \textit{Long} task suite, a larger pool of 208 trajectories is employed to address its more challenging, long-horizon nature. In contrast, for the $\pi_{0.5}$ model, benefiting from a superior pre-trained checkpoint and training configurations, we leverage only 40 trajectories to provide a unified few-shot SFT checkpoint across all task suites.

\textbf{RL Procedure.}
In RL, the VLA model receives a multi-modal input state comprising: an agent-view and a wrist-view, natural language guidance, the robot end effector pose, and the gripper state. The model outputs an action to interact with the LIBERO environment, which provides a binary reward of 1 for successful task completion and 0 otherwise.

\textbf{Experiments.}
We benchmark the performance of $\pi_{\texttt{RL}}$, which fine-tunes the few-shot SFT $\pi_0$ and $\pi_{0.5}$ models with Flow-Noise and Flow-SDE, against several state-of-the-art VLAs trained on the entire LIBERO dataset, including Octo, OpenVLA, OpenVLA-OFT, $\pi_{\text{fast}}$ \citep{pertsch2025fast}, $\pi_{0}$, and $\pi_{0.5}$. We conduct experiments on four LIBERO task suites and report performance as the success rate across all 500 initial states (10 sub-tasks $\times$ 50 states each).

\textbf{Analysis.} 
For the few-shot $\pi_0$ model, the SFT baseline performs poorly, with an average success rate of only 57.6\%, indicating that the model struggles with limited demonstration data. $\pi_{\texttt{RL}}$ substantially boosts performance, with Flow-SDE and Flow-Noise reaching 96.1\% and 97.6\%, and surpassing the full-dataset $\pi_0$ SFT baseline of 94.2\%.

While the $\pi_{0.5}$ few-shot SFT baseline achieves a decent average performance of 77.1\%, it struggles with the challenging \textit{Long} task suite, scoring only 43.9\%. Our proposed $\pi_{\texttt{RL}}$ rectifies this deficiency, boosting the \textit{Long} task success rate from 43.9\% to 94.0\%, constituting a 50.1\% improvement. Notably, despite using only a single trajectory for SFT, $\pi_{\texttt{RL}}$ reaches 98.3\% final performance, surpassing the 96.9\% full-dataset SFT model.

\begin{table*}[t]
    \centering
    \caption{Evaluation results on the CALVIN benchmark (Scene D), reporting the average completed subtasks and success rates for task sequences of length 1 to 5.}
    \label{tab:comparison_calvin}
    \begin{tabular*}{\textwidth}{@{\extracolsep{\fill}}llccccccc@{}} 
        \toprule
        \multicolumn{2}{l}{\multirow{2}{*}{\textbf{Methods}}} & \multicolumn{7}{c}{\textbf{CALVIN-D}} \\ 
        \cmidrule(lr){3-9}
        \multicolumn{2}{c}{} & \textbf{Len-1} & \textbf{Len-2} & \textbf{Len-3} & \textbf{Len-4} & \textbf{Len-5} & \textbf{Avg.} & \textbf{$\Delta$ Avg.} \\ 
        \midrule
        \hspace{0.38em} \multirow{3}{*}{$\pi_0$} &
            \hspace{0.38em} SFT & 94.7 & 84.9 & 74.3 & 65.2 & 57.5 & 3.766 & --- \\
        & \hspace{0.38em} Flow-SDE & 96.4 & 88.0 & 77.5 & 70.8 & 61.7 & 3.944 & \deltaimp{0.178} \\
        & \hspace{0.38em} Flow-Noise & 96.9 & 88.8 & 78.0 & 68.3 & 59.9 & 3.919 & \deltaimp{0.153} \\
        \midrule
        \hspace{0.38em} \multirow{3}{*}{$\pi_{0.5}$} &
            \hspace{0.38em} SFT & 92.7 & 84.3 & 76.7 & 68.8 & 61.3 & 3.838 & --- \\
        & \hspace{0.38em} Flow-SDE & \textbf{99.7} & \textbf{98.2} & \textbf{95.8} & \textbf{91.0} & \textbf{87.0} & \textbf{4.717} & \deltaimp{\textbf{0.879}} \\
        & \hspace{0.38em} Flow-Noise & 99.6 & 97.6 & 93.9 & 89.6 & 84.5 & 4.652 & \deltaimp{0.814} \\
        \bottomrule
    \end{tabular*}
\end{table*}
\begin{table*}[t]
    \centering
    \caption{Evaluation results on the MetaWorld MT50 benchmark.}
    \label{tab:comparison_metaworld}
    \begin{tabular*}{\textwidth}{@{\extracolsep{\fill}}llcccccc@{}} 
        \toprule
        \multicolumn{2}{l}{\multirow{2}{*}{\textbf{Methods}}} & \multicolumn{6}{c}{\textbf{MetaWorld-MT50}} \\ 
        \cmidrule(lr){3-8}
        \multicolumn{2}{c}{} & \textbf{Easy} & \textbf{Medium} & \textbf{Hard} & \textbf{Very Hard} & \textbf{Avg.} & \textbf{$\Delta$ Avg.} \\ 
        \midrule
        \multicolumn{2}{l}{Diffusion Policy} & 23.1 & 10.7 & 1.9 & 6.1 & 10.5 & --- \\
        \multicolumn{2}{l}{TinyVLA} & 77.6 & 21.5 & 11.4 & 15.8 & 31.6 & --- \\
        \multicolumn{2}{l}{SmolVLA} & 87.1 & 51.8 & 70.0 & 64.0 & 68.2 & --- \\
        \midrule
        \hspace{0.38em} \multirow{3}{*}{$\pi_0$} &
            \hspace{0.38em} SFT & 77.9 & 51.8 & 53.3 & 20.0 & 50.8 & --- \\
        & \hspace{0.38em} Flow-SDE & \textbf{92.1} & 74.6 & 61.7 & 84.0 & 78.1 & \deltaimp{27.3} \\
        & \hspace{0.38em} Flow-Noise & 91.1 & \textbf{81.8} & \textbf{78.3} & \textbf{92.0} & \textbf{85.8} & \deltaimp{\textbf{35.0}} \\
        \midrule
        \hspace{0.38em} \multirow{3}{*}{$\pi_{0.5}$} &
            \hspace{0.38em} SFT & 68.2 & 37.3 & 41.7 & 28.0 & 43.8 & --- \\
        & \hspace{0.38em} Flow-SDE & 86.4 & 55.5 & 75.0 & 66.0 & 70.7 & \deltaimp{26.9} \\
        & \hspace{0.38em} Flow-Noise & 86.8 & 58.1 & 63.3 & 56.0 & 66.1 & \deltaimp{22.3} \\
        \bottomrule
    \end{tabular*}
\end{table*}

\subsection{ManiSkill Benchmark} 
\textbf{SFT Procedure.}
In the ManiSkill benchmark, the policy is required to pick from 16 object types and place them onto 17 receptacles across 16 unique table scenes, yielding 4,352 unique task combinations. Given the high complexity of this setting, the SFT dataset consists of 16,384 episodes synthesized using the MPLib motion planning suite \citep{Guo_MPlib}. To reinforce the concept of motion completion, 15 additional frames are appended to the end of each trajectory.

\textbf{RL Procedure.}
In RL, the VLA model receives a third-person RGB image, a concise language instruction, and the current joint proprioception. The environment provides a structured reward signal: 1.0 for correct object placement and an auxiliary 0.1 reward for successful gripper-object attachment, intended to encourage stable manipulation and mitigate undesired behaviors such as impulsive throwing.

\textbf{Experiments.} Following the RL4VLA experimental protocol, we conduct RL training on a comprehensive set of 4,352 task combinations and record the performance as the aggregate success rate across these tasks.

\textbf{Analysis.} As detailed in \cref{tab:ind_ood_comparison}, $\pi_{\texttt{RL}}$ significantly boosts performance in the training environment. Specifically, the success rate of $\pi_0$ increases from 38.4\% to 77.8\%, while $\pi_{0.5}$ improves from 40.1\% to 90.9\%. These gains underscore the efficacy of RL in complex settings.

\subsection{CALVIN Benchmark} 
\textbf{SFT Procedure.} We conduct SFT on the CALVIN ABC dataset\footnote{https://huggingface.co/datasets/InternRobotics/InternData-Calvin\_ABC}, which comprises approximately 24 hours of unstructured "play data" across three distinct environments (A, B, and C). This dataset includes over 20,000 language-labeled trajectories covering 34 unique manipulation tasks. 

\textbf{RL Procedure.} Each episode consists of a sequence of five randomly sampled subtasks to be completed in succession without environment resets. The reward signal is defined at the subtask level, where the model receives a sparse binary reward of 1.0 for each successfully executed subtask and 0.0 otherwise.

\begin{table*}[htbp]
\centering
\caption{Specific generalization evaluation results in the ManiSkill OOD setting.}
\label{tab:maniskill_ood_results}
\resizebox{\textwidth}{!}{%
\begin{tabular}{@{}llcccccc@{}}
\toprule
\multirow{2}{*}{\textbf{Environment}} & \multirow{2}{*}{\textbf{Variation-Version-Type}} & \multirow{2}{*}{\textbf{$\pi_0$-SFT}} & \textbf{$\pi_0$-RL} & \textbf{$\pi_0$-RL} & \multirow{2}{*}{\textbf{$\pi_{0.5}$-SFT}} & \textbf{$\pi_{0.5}$-RL} & \textbf{$\pi_{0.5}$-RL} \\
 &  &  & \textbf{Flow-SDE} & \textbf{Flow-Noise} &  & \textbf{Flow-SDE} & \textbf{Flow-Noise} \\ \midrule
In distribution & Main-v3-train & 38.4 & 78.8 & 77.8 & 40.1 & \textbf{90.9} & 89.7 \\ \midrule
\multirow{6}{*}{\makecell[l]{Visual-Language\\Variations}} & Instruct-v1-test & 30.1 & 64.6 & 66.5 & 46.6 & 77.0 & \textbf{85.7} \\
 & VisionImage-v1-test & 38.3 & 68.8 & 71.7 & 46.2 & 78.8 & \textbf{83.1} \\
 & VisionTexture03-v1-test & 35.1 & 66.0 & 66.8 & 36.7 & 69.6 & \textbf{75.0} \\
 & VisionTexture05-v1-test & 31.0 & 55.8 & 60.5 & 32.7 & 58.0 & \textbf{62.2} \\
 & VisionWhole03-v1-test & 35.4 & 62.4 & 69.0 & 40.1 & 69.6 & \textbf{71.6} \\
 & VisionWhole05-v1-test & 28.5 & 49.0 & 53.9 & 30.7 & 55.0 & \textbf{57.0} \\ \midrule
\multirow{4}{*}{\makecell[l]{Semantic Reasoning\\(object/receptacle\\confounders)}} & MultiCarrot-v1-test & 7.8 & 28.2 & 23.0 & 16.7 & 36.8 & \textbf{38.2} \\
 & MultiCarrot-v1-train & 12.5 & 36.5 & 31.8 & 28.2 & 49.5 & \textbf{50.1} \\
 & MultiPlate-v1-test & 5.0 & 16.4 & 18.3 & 11.8 & \textbf{29.4} & 28.3 \\
 & MultiPlate-v1-train & 7.3 & 20.5 & 19.6 & 9.7 & 22.3 & \textbf{25.4} \\ \midrule
\multirow{2}{*}{Action Execution} & PositionChangeTo-v1-test & 9.6 & 17.4 & 10.9 & 13.5 & 36.2 & \textbf{54.7} \\
 & Position-v1-test & 16.9 & 45.6 & 37.5 & 31.2 & 54.5 & \textbf{55.0} \\ \bottomrule
\end{tabular}%
}
\end{table*}

\textbf{Experiments.} 
We evaluate the performance of $\pi_{\texttt{RL}}$ in Scene D over 1,000 episodes. Following the standard CALVIN evaluation protocol, we report two key metrics: (1) the success rate for task sequences of increasing lengths, namely \textit{Len-1} to \textit{Len-5}. (2) the average number of completed sub-tasks, denoted as \textit{Avg.} per episode. 

\textbf{Analysis.} 
As detailed in \cref{tab:comparison_calvin}, $\pi_{\texttt{RL}}$ yields substantial performance gains in long-horizon sequential execution, particularly with the $\pi_{0.5}$ variant. The SFT models inherently struggle with compounding errors across sequential tasks, with $\pi_{0.5}$ only achieving a 61.3\% success rate on Len-5 sequences. RL effectively mitigates this issue with the average completed sub-tasks of Flow-SDE increasing from 3.838 to 4.717, and its Len-5 success rate surges to 87.0\%. 

Notably, the performance gap between SFT and RL widens significantly as the sequence length increases. For $\pi_{0.5}$, while Flow-SDE shows a modest 7.0\% improvement over SFT in Len-1 tasks, the gap expands to an impressive 25.7\% in the most challenging Len-5 sequences.

\subsection{MetaWorld Benchmark} 
\textbf{SFT Procedure.}
We perform SFT on the $\pi_0$ and $\pi_{0.5}$ models using the official dataset\footnote{https://huggingface.co/datasets/lerobot/metaworld\_mt50}, which consists of 2500 trajectories across 50 different manipulation tasks.

\textbf{RL Procedure.}
During the RL procedure, the VLA model processes a multi-modal input comprising a RGB agent-view image, language guidance, the robot's end-effector position, and its gripper state. Based on this input, the model outputs an action to interact with the environment, which in turn provides a sparse reward: 1 for successful task completion and 0 otherwise.

\textbf{Experiments.}
We benchmark the performance of $\pi_{\texttt{RL}}$ against Diffusion Policy \citep{diffusion_policy}, TinyVLA \citep{wen2025tinyvla}, and SmolVLA \citep{shukor2025smolvla}. For the performance evaluation, we follow the setup from SmolVLA, \ie, classifying 50 tasks into \textit{easy}, \textit{medium}, \textit{hard}, and \textit{very hard} four categories according to their difficulties.

\textbf{Analysis.}
As detailed in \cref{tab:comparison_metaworld}, RL fine-tuning substantially boosts performance. The $\pi_0$ and $\pi_{0.5}$ models achieve average success rates of 85.8\% and 70.7\%, respectively. This marks a significant improvement over their SFT-only counterparts and surpasses the SmolVLA baseline of 68.2\%, confirming that RL can effectively enhance model capabilities across a diverse range of manipulation task types.

\section{Out-of-Distribution RL Evaluation}
\label{appendix:ood_experiments}
While previous experiments demonstrate significant RL-driven improvements in the ID domain, this section evaluates OOD generalization. As LIBERO lacks a dedicated interface for OOD testing, we utilize the ManiSkill, CALVIN, and MetaWorld benchmarks to investigate \textit{whether the RL-driven improvements represent genuine skill acquisition that scales to novel settings, or merely reflect the exploitation of environment-specific biases.}

\begin{table*}[t]
\centering
\caption{Evaluation results on the SIMPLER benchmark for $\pi_0$ and $\pi_{0.5}$ with Flow-Noise method.}
\label{tab:comparison_simpler}
\begin{tabular*}{\textwidth}{@{\extracolsep{\fill}}llccccc}
\toprule
\multicolumn{2}{l}{\multirow{2}{*}{\textbf{Model}}} & \multicolumn{5}{c}{\textbf{SIMPLER}} \\
\cmidrule(lr){3-7}
\multicolumn{2}{l}{} & \textbf{Carrot} & \textbf{Eggplant} & \textbf{Spoon} & \textbf{Cube} & \textbf{Avg.} \\
\midrule
\hspace{0.38em} \multirow{3}{*}{$\pi_0$}
    & \hspace{0.38em} SFT      & 82.7 & 87.5 & 61.7 & 37.1 & 67.2 \\
    & \hspace{0.38em} +RL      & \textbf{95.7} & 96.7 & \textbf{91.6} & \textbf{63.0} & \textbf{86.7} \\
    & \hspace{0.38em} $\Delta$ & \deltaimp{13.0} & \deltaimp{9.2} & \deltaimp{29.9} & \deltaimp{25.9} & \deltaimp{19.5} \\
\midrule
\hspace{0.38em} \multirow{3}{*}{$\pi_{0.5}$}
    & \hspace{0.38em} SFT      & 70.6 & 91.9 & 43.5 & 31.0 & 59.2 \\
    & \hspace{0.38em} +RL      & 82.0 & \textbf{98.2} & 82.8 & 53.3 & 79.1 \\
    & \hspace{0.38em} $\Delta$ & \deltaimp{11.4} & \deltaimp{6.3} & \deltaimp{39.3} & \deltaimp{22.3} & \deltaimp{19.9} \\
\bottomrule
\end{tabular*}
\end{table*}

\begin{figure*}
    \centering
    \includegraphics[width=1\linewidth]{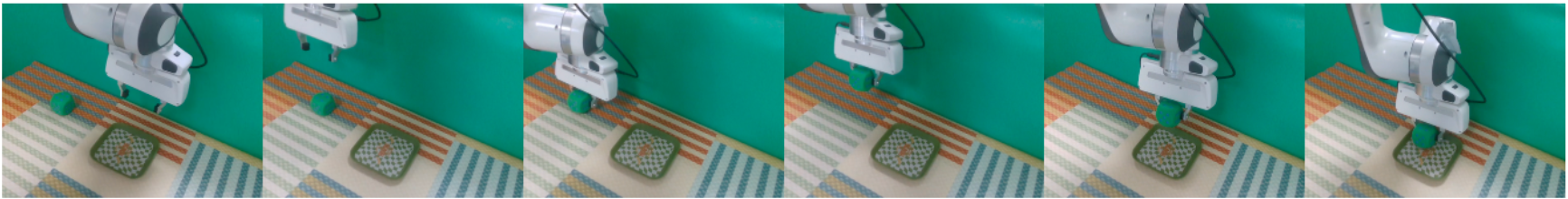}
    \caption{Real-world deployment of an RL refined policy performing a pick and place task.}
    \label{fig:realrobot}
\end{figure*}

\subsection{ManiSkill}
\textbf{Setup.} Following RL4VLA, we evaluate the model's generalization across three challenging OOD scenarios: (1) \textit{Vision}, challenging the model with novel backgrounds and textures; (2) \textit{Semantics}, probing comprehension with unseen objects, varied instructions, and confounding elements like extra objects or receptacles; (3) \textit{Execution}, assessing robustness against varied initial states, unseen robot poses, and dynamic disturbances. 

\textbf{Results.} In the OOD scenarios detailed in \cref{tab:ind_ood_comparison,tab:maniskill_ood_results}, we observe that the $\pi_{0}$-SFT model demonstrates strong generalization for visual information. This can be attributed to the robust foundation of its VLM, which allows it to handle visual disturbances better.

However, the semantic performance of $\pi_0$ drops dramatically. This degradation is less pronounced when switching to the $\pi_{0.5}$ baseline, a benefit likely stemming from the knowledge generalization of the pre-trained $\pi_{0.5}$ model. Regarding action execution, $\pi_0$ exhibits a larger performance drop than $\pi_{0.5}$. We hypothesize that this discrepancy arises from the inclusion of joint angle states as input in $\pi_0$, leading to severe overfitting in the control task. In contrast, $\pi_{0.5}$ omits these inputs, thereby avoiding the same degree of performance degradation.

As for the RL training, although the performance improvements in OOD scenarios are lower than those in IND settings, the proportional improvements achieved are notably comparable. As indicated in \cref{tab:ind_ood_comparison}, for the $\pi_{0.5}$ model, Flow-SDE enhances the IND success rate by 126.7\%, while the OOD similarly increases by 102.3\%. This consistency in relative gains indicates that RL-driven optimization promotes the acquisition of generalized action representations rather than merely overfitting the training environment, thus preserving efficacy under distribution shifts.

Nevertheless, we observe a performance gap in relative improvement between the \textit{Vision} OOD tasks and the IND domain. Specifically, for $\pi_{0.5}$, the 73.9\% gain in visual generalization trails the 126.7\% increase observed in the training environment. This discrepancy likely stems from freezing the VLM backbone during the RL stage for computational efficiency, which restricts the model's ability to adapt its visual grounding features to novel textures and backgrounds. 

\subsection{CALVIN}
\textbf{Setup.} We evaluate environmental and visual OOD robustness based on the ABC $\to$ D protocol in CALVIN. Under this setting, the model is trained on ABC environments and evaluated in a zero-shot manner on Scene D. Scene D introduces significant distribution shifts in terms of visual textures, lighting conditions, and spatial layouts, effectively assessing the agent's ability to transfer skills to an unfamiliar physical environment.

\textbf{Results.} 
Under identical D$\to$ D training settings, the RL finetuned policy in the ABC environment reaches a 79.1\% success rate in the OOD scene D, improving over the 61.3\% SFT baseline as shown in \cref{fig:ood_overall}. Aligned with the findings in ManiSkill, this suggests that ID gains can be transferred to OOD settings characterized by visual variations.

\subsection{MetaWorld}
\textbf{Setup.} We utilize the ML45 benchmark from MetaWorld to evaluate the task-level generalization. This setup consists of 50 distinct robotic manipulation tasks: the agent is trained on 45 base tasks and subsequently evaluated on 5 held-out, unseen tasks, which require the model to generalize its learned manipulation primitives to entirely novel task objectives and workspace configurations.

\textbf{Results.} 
As evidenced in the \cref{fig:ood_overall}, while success rates show consistent gains within the ID domain, OOD performance is characterized by persistent oscillation throughout the training process. This instability indicates that RL, in its current form, struggles to foster stable cross-category generalization. 

Nevertheless, the model retains the OOD skills learned during the SFT phase throughout the RL training process. This highlights a significant advantage over standard SFT, which often causes the model to overfit on expert demonstrations and lose its broader capabilities \cite{li2025simplevla}. Unlike standard SFT, RL enables ID performance gains while preserving the general knowledge, which indicates that RL provides a more balanced optimization that maintains OOD robustness without catastrophic forgetting.

\begin{table*}[t]
\centering
\caption{Comparison of the PPO and GRPO with Flow-SDE on the LIBERO.}
\label{tab:ppo_grpo_comparison}
\begin{tabular*}{\textwidth}{@{\extracolsep{\fill}}llcccccl} 
\toprule
\multicolumn{2}{l}{\multirow{2}{*}{\textbf{Model}}} & \multicolumn{6}{c}{\textbf{LIBERO}} \\ 
\cmidrule(lr){3-8}
\multicolumn{2}{l}{} & \textbf{Spatial} & \textbf{Object} & \textbf{Goal} & \textbf{Long} & \textbf{Avg.} & \textbf{$\Delta$ Avg.} \\ 
\midrule
\hspace{0.38em} \multirow{3}{*}{$\pi_0$} 
    & \hspace{0.38em} SFT       & 65.3 & 64.4 & 49.8 & 51.2 & 57.6 & --- \\ 
    & \hspace{0.38em} +GRPO & 97.8 & 97.8 & 83.2 & 81.4 & 90.0 & \deltaimp{32.4} \\ 
    & \hspace{0.38em} +PPO   & 98.4 & 99.4 & 96.2 & 90.2 & 96.0 & \deltaimp{\textbf{38.4}} \\ 
\midrule
\hspace{0.38em} \multirow{3}{*}{$\pi_{0.5}$} 
    & \hspace{0.38em} SFT       & 84.6 & 95.4 & 84.6 & 43.9 & 77.1 & --- \\ 
    & \hspace{0.38em} +GRPO & 97.4 & 99.8 & 91.2 & 77.6 & 91.5 & \deltaimp{14.4} \\ 
    & \hspace{0.38em} +PPO   & \textbf{99.6} & \textbf{100} & \textbf{98.8} & \textbf{93.0} & \textbf{97.9} & \deltaimp{20.8} \\ 
\bottomrule
\end{tabular*}
\end{table*}

\subsection{Summary}
In conclusion, our OOD evaluation demonstrates that \textit{RL enhances performance for similar tasks but fails to generalize effectively to novel task objectives}. 

Specifically, RL training effectively enhances robustness against low-level variations such as the visual and execution shifts observed in ManiSkill and CALVIN. This indicates that the model acquires generalized action representations rather than merely overfitting to the training environment.

Regarding high-level generalization on MetaWorld, the model successfully retains the OOD skills inherited from the SFT phase, demonstrating that RL avoids the catastrophic forgetting and overfitting typical of standard imitation learning. However, transferring its performance gains to entirely novel task objectives remains a significant challenge.

\section{Case Studies: Single-Task RL Training}
\label{appendix:case_study}
While the preceding experiments focused on performance across multi-task benchmarks, this section investigates single-task scenarios where the VLA is trained to master a specific task within a relatively static environment. Specifically, we evaluate our approach on the \textbf{SIMPLER} \cite{SIMPLER} and a \textbf{Real2Sim2Real} environment. 

\subsection{SIMPLER}
\textbf{Setup.} In SIMPLER, the experimental setup comprises an 8-DoF WidowX-250S arm evaluated on four standard tasks: (1) \texttt{Spoon}: placing a spoon on a cloth. (2) \texttt{Carrot}: placing a carrot on a plate. (3) \texttt{Eggplant}: placing an eggplant in a basket. (4) \texttt{Cube}: stacking a cube. For the SFT stage, we employ a curated dataset in which each task is trained with 144 demonstration episodes.

\textbf{Analysis.} As detailed in \cref{tab:comparison_simpler}, $\pi_{\texttt{RL}}$ increases the average success rate of the $\pi_0$ model from 67.2\% to 86.7\%, with three tasks (\texttt{carrot}, \texttt{eggplant}, and \texttt{spoon}) exceeding 90\% success. 

\begin{figure*}[t!] 
    \centering
    \begin{subfigure}{0.32\textwidth}
        \centering
        \includegraphics[width=\linewidth]{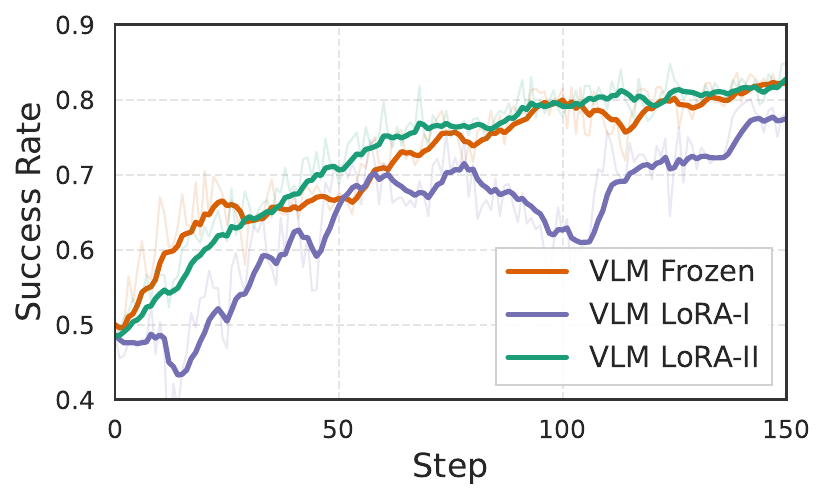}
        \caption{Eval}
    \end{subfigure}%
    \hspace{5em} 
    \begin{subfigure}{0.32\textwidth}
        \centering
        \includegraphics[width=\linewidth]{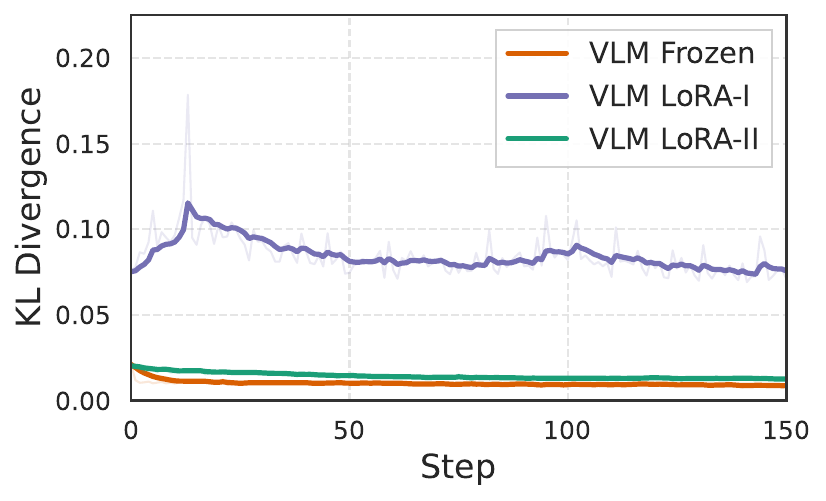}
        \caption{KL Divergence}
    \end{subfigure}%
    \caption{
    Ablation study on the effectiveness of VLM during RL. 
    }
    \label{fig:lora_libero}
\end{figure*}
\begin{figure*}[t!] 
    \centering
    \begin{subfigure}{0.32\textwidth}
        \centering
        \includegraphics[width=\linewidth]{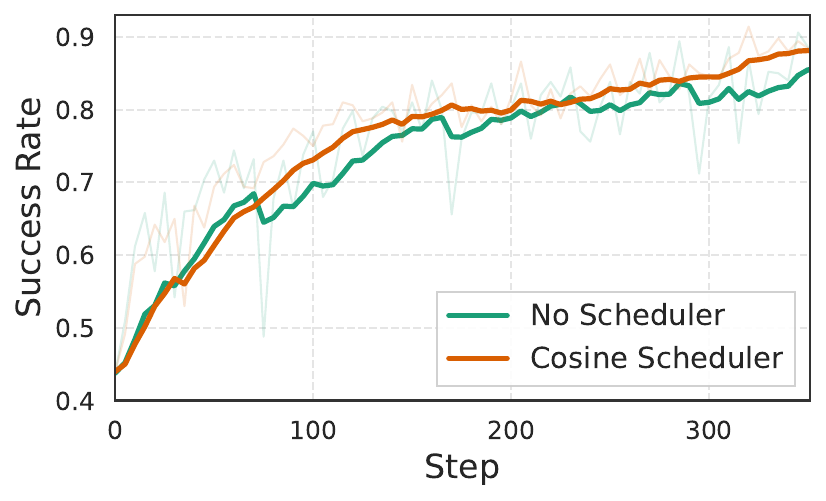}
        \caption{Eval}
    \end{subfigure}%
    \hspace{5em} 
    \begin{subfigure}{0.32\textwidth}
        \centering
        \includegraphics[width=\linewidth]{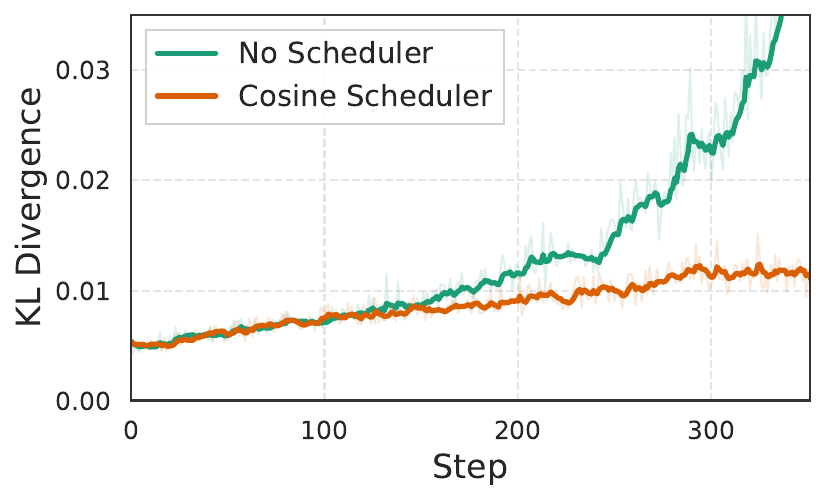}
        \caption{KL Divergence}
    \end{subfigure}%
    \caption{
    Ablation study on the learning rate scheduler. 
    }
    \label{fig:lr_scheduler}
\end{figure*}

\subsection{Real2Sim2Real}
While the SIMPLER benchmark demonstrates predictive correlation between simulation and real-world performance, a pronounced visual domain shift remains to be solved. To this end, we leverage recent Real2Sim2Real methodologies \cite{fan2025twinaligner, jiang2025gsworld} to construct a high-fidelity simulation environment with ManiSkill for rigid body dynamics and Gaussian Splatting \cite{3dgs} for photorealistic rendering.

\textbf{Setup.} Our hardware platform comprises a Franka Panda robotic arm and an Intel RealSense D435 camera serving as the primary visual sensor. We perform manual calibration by aligning simulated viewpoints with real-world camera perspectives to synchronize their extrinsic matrices. As illustrated in \cref{fig:top_figure}, the photorealistic textures and color profiles in our simulator closely mirror the physical environment, effectively minimizing the visual domain shift from simulation to reality.

\textbf{Results.} Following the experimental protocol established in the ManiSkill benchmarks, we initially collect 20 expert trajectories via a motion planner for few-shot SFT, which is subsequently optimized through RL over 100 training iterations. We deploy the RL fine-tuned policies in the real world in a zero-shot manner. Empirical results indicate that while the SFT baseline fails to complete the task, the RL-tuned policy achieves a 40\% success rate. A representative successful episode is visualized in \cref{fig:realrobot}.

\section{Ablation Details}
\label{appendix:ablation}
\subsection{RL algorithms}
Given the significant performance gains from PPO on the LIBERO benchmark, we also investigated the effectiveness of GRPO \citep{shao2024grpo}, another widely used policy gradient method applied in VLA+RL training \citep{li2025simplevla}. We compare the performance of PPO and GRPO on both the $\pi_0$ and $\pi_{0.5}$ models, with results denoted in \cref{tab:ppo_grpo_comparison}.


\textbf{Conclusion.} To sum up, our findings highlight a critical trade-off: \textit{parameters tailored for rollout success may adversely impact training stability, ultimately constraining the performance ceiling of RL.} Therefore, careful parameter tuning is required to achieve a synergy between high-quality rollouts and stable policy convergence.

\begin{figure*}[t!] 
    \centering
    \begin{subfigure}{0.32\textwidth}
        \centering
        \includegraphics[width=\linewidth]{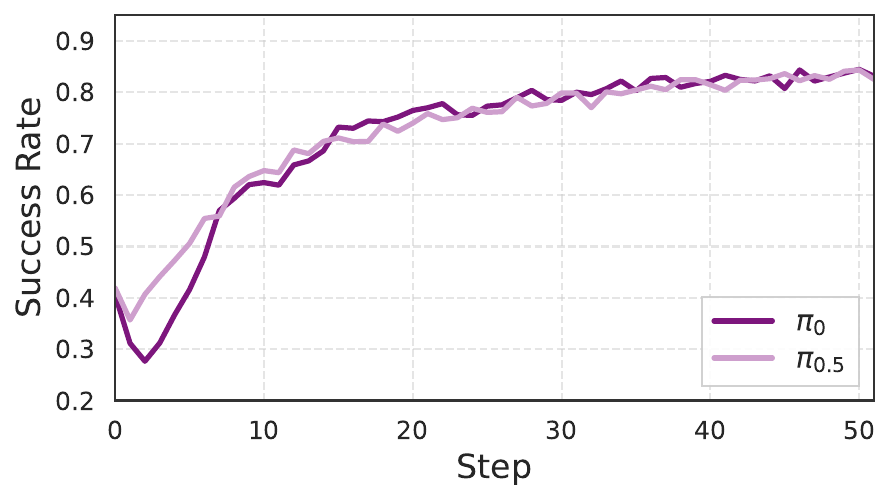}
        \caption{Eval}
    \end{subfigure}%
    \hspace{5em} 
    \begin{subfigure}{0.32\textwidth}
        \centering
        \includegraphics[width=\linewidth]{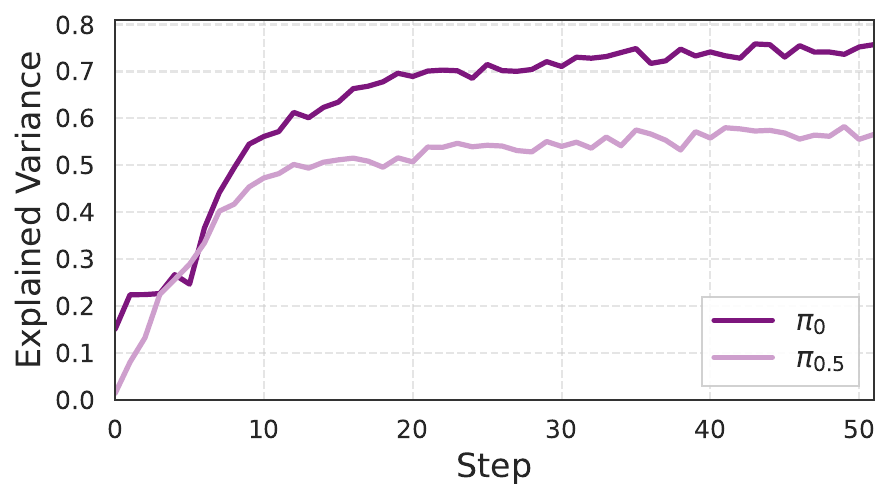}
        \caption{Explained Variance}
    \end{subfigure}%
    \caption{Training curves in ManiSkill.}
    \label{fig:critic_warm}
\end{figure*}
\begin{figure*}[t!] 
    \centering
        \centering
        \includegraphics[width=0.4\linewidth]{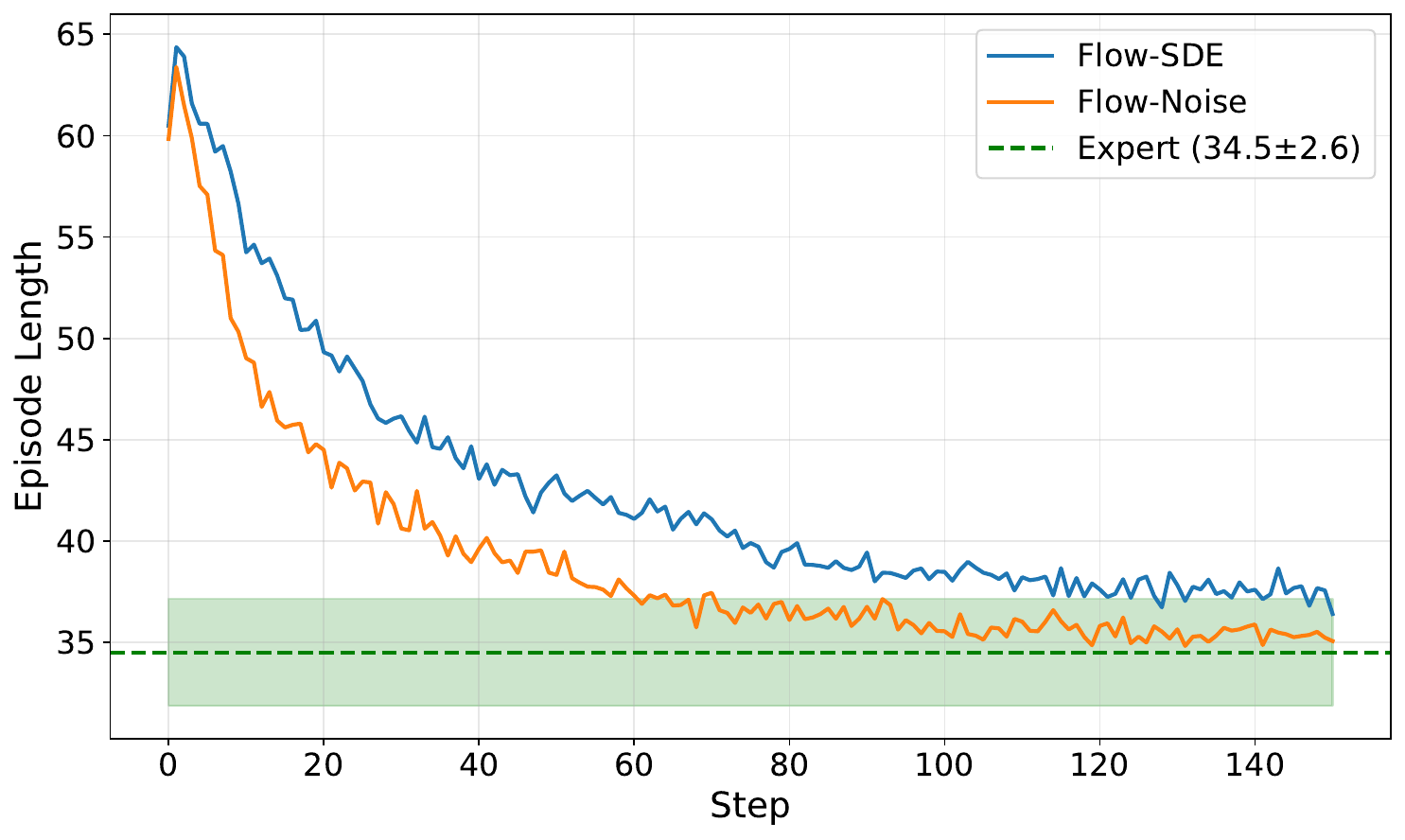}
\caption{Episode length: $\pi_{0.5}$ RL training in ManiSkill.}
    \label{fig:eps_len}
\end{figure*}

\subsection{VLM Fine-tuning Analysis}
In our previous experiments, the VLM is frozen, and the optimization is confined exclusively to the action expert during RL. In this subsection, we aim to investigate the role of the VLM during RL. Specifically, we employ Low-Rank Adaptation (LoRA)~\citep{hu2022lora} for the VLM, facilitating its joint optimization with the action expert. We set the LoRA rank to $r=32$ and the scaling parameter to $\alpha=32$, while the action expert remains fully trainable. 

We conduct experiments with the $\pi_0$ model with Flow-SDE on the LIBERO-Long benchmark, comparing three distinct configurations: 
\begin{itemize}
    \item \textbf{VLM Frozen}: $5e^{-6}$ learning rate, 4 updates/epoch.
    \item \textbf{VLM LoRA-I}: $5e^{-6}$ learning rate, 4 updates/epoch.
    \item \textbf{VLM LoRA-II}: $1e^{-6}$ learning rate, 2 updates/epoch. 
\end{itemize}

As presented in \cref{fig:lora_libero}, the VLM LoRA-II configuration achieves a learning trajectory comparable to the VLM frozen baseline. This empirical observation yields two critical inferences: 
\textbf{(1)} The benefit of fine-tuning the VLM on the LIBERO benchmark is not evident. We conjecture the limited performance gain owing to the limited scene variability within LIBERO, for which the pretrained VLM representations are already sufficiently robust. \textbf{(2)} Fine-tuning VLM together with the action expert requires a more conservative optimization configuration for training stability. 

\section{Insights from Large-Scale RL Training}
\label{appendix:large_scale_rl_trianing}
In this section, we elaborate on some empirical insights we gained during RL training.

\textbf{Hyperparameters.}
According to the hyperparameters ablation detailed in \cref{subsec:hyperparameters}, the performance disparity between the train and eval performance of the initial SFT checkpoint warrants close attention. If this disparity is significant, we recommend either reducing the noise magnitude or increasing the number of denoising steps to mitigate the performance loss when shifting from deterministic to stochastic execution. Furthermore, as previously established, lower noise levels yield larger gradients, requiring a smaller learning rate to maintain training stability.

We also observed that when train performance improves steadily while eval performance oscillates, increasing the number of denoising steps can help alleviate this, benefiting from reduced divergence in the action distributions between the deterministic and stochastic action generation processes. Regarding the action chunk, we empirically found that long-horizon tasks benefit from larger chunk sizes. For instance, we set the chunk size to 10 for LIBERO-Long and 5 for the other sub-tasks. 

\textbf{Training.}
In our $\pi_{0.5}$ experiments on the LIBERO-Long benchmark, we observed that the Kullback–Leibler (KL) divergence metric increased steadily throughout training, potentially leading to instability. We mitigated this issue by implementing a learning rate scheduler with cosine annealing. As demonstrated in \cref{fig:lr_scheduler}, this scheduler effectively prevents the KL divergence from escalating, thereby stabilizing the training process. 

\textbf{Critic.} In our ManiSkill experiments, we observe that policy evaluation performance exhibits an initial dip before improving for both $\pi_0$ and $\pi_{0.5}$ models, as shown in \cref{fig:critic_warm}. We attribute this transient degradation to the critic providing inaccurate signals during its warm-up phase. The subsequent eval improvement correlates directly with the critic's value estimations stabilizing, as evidenced by the rising explained variance.

\textbf{Temporal Efficiency.}
We also study how the rollout of RL in a physical simulator helps shape the policy to achieve expert-level temporal efficiency. 
We analyze the expert motion planning data used for SFT and then tracked the average episode lengths during the RL training of the $\pi_{0.5}$ model in ManiSkill. 
As shown in \cref{fig:eps_len}, the SFT-initialized policy exhibits significantly longer episodes due to execution errors. In contrast, $\pi_{0.5}$ achieves episode lengths that converge to the expert range after RL training, demonstrating a substantial improvement in temporal efficiency.

We attribute this convergence to two factors: \textbf{(1)} RL enhances the policy’s error-correction capabilities, allowing it to recover from execution failures. \textbf{(2)} Our partial reset mechanism incentivizes temporal efficiency through discounted rewards, as faster task completion enables the agent to trigger more resets and accumulate higher total rewards within each update cycle.

\begin{figure*}[!t]
    \centering
    \includegraphics[width=0.7\linewidth]{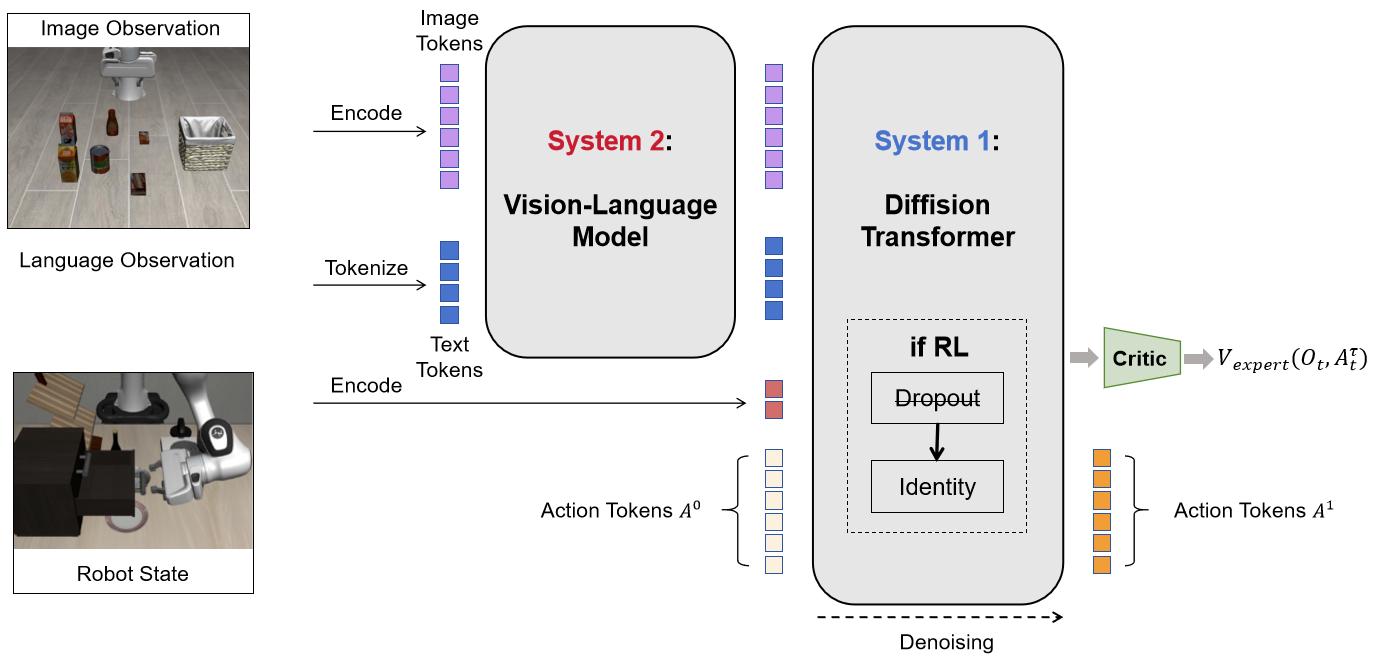}
    \caption{Illustration for the architecture of GR00T-N1.5.}
    \label{fig:gr00t_arch}
\end{figure*}

\begin{table*}[t]
\centering
\caption{Results of Finetuning GR00T using PPO with Flow-SDE on the LIBERO.}
\label{tab:gr00t_result}
\begin{tabular*}{\textwidth}{@{\extracolsep{\fill}}llcccccl} 
\toprule
\multicolumn{2}{l}{\multirow{2}{*}{\textbf{Model}}} & \multicolumn{6}{c}{\textbf{LIBERO}} \\ 
\cmidrule(lr){3-8}
\multicolumn{2}{l}{} & \textbf{Spatial} & \textbf{Object} & \textbf{Goal} & \textbf{Long} & \textbf{Avg.} & \textbf{$\Delta$ Avg.} \\ 
\midrule
\hspace{0.38em} \multirow{2}{*}{GR00T} 
    & \hspace{0.38em} SFT       & 41.4 & 58.6 & 48.2 & 61.9 & 52.5 & --- \\ 
    & \hspace{0.38em} +PPO   & 92.5 & 96.2 & 84.3 & 86.6 & 89.9 & \deltaimp{\textbf{37.4}} \\ 
\bottomrule
\end{tabular*}
\end{table*}

\section{RL for GR00T N1.5}
\label{appendix:gr00t}
\subsection{Setup}

\textbf{GR00T N1.5.}
We conduct additional experiments based on the GR00T N1.5 model \cite{bjorck2025gr00t}, which is a foundation model tailored for generalist humanoid robot reasoning and manipulation. 
The architecture integrates an Eagle 2.5 VLM \citep{chen2025eagle}, optimized for spatial grounding and physical reasoning, with a Diffusion Transformer head \citep{dit} for action denoising. It facilitates multi-embodiment compatibility through specialized heads, supporting configurations such as humanoids with dexterous hands or grippers, as well as single-arm manipulators. 

Regarding the critic implementation, we estimate value functions across the entire denoising trajectory by integrating the critic network directly with the action head. The complete framework is illustrated in \cref{fig:gr00t_arch}.

\textbf{Benchmark.} We evaluate the model performance of GR00T across four manipulation task suites in LIBERO: \textit{Spatial}, \textit{Object}, \textit{Goal}, and \textit{Long}.

\textbf{Implementation Details.}
Similar to the $\pi_0$ implementation, we initiate our process with SFT on expert demonstrations. For the SFT stage,  we fine-tune the entire model following the official setting. 
In the subsequent RL stage, we exclusively fine-tune the action expert model while keeping the vision-language model parameters fixed. 

A crucial methodological refinement in our RL pipeline is the replacement of dropout layers in the expert model with identity layers. Dropout is widely recognized to induce instability during online RL training. Specifically, it introduces non-deterministic perturbations to the effective policy, shifting the standard probability ratio update from:
\begin{equation}
\rho_t(\theta) = \frac{\pi_{\theta_{\text{new}}}(a_t|s_t)}{\pi_{\theta_{\text{old}}}(a_t|s_t)}
\end{equation}
to a highly unstable form:
\begin{equation}
\rho_t(\theta) = \frac{\pi_{\alpha_{\text{new}}}(a_t|s_t)}{\pi_{\theta_{\text{old}}}(a_t|s_t)},
\end{equation}
where $\rho_t(\theta)$ denotes the probability ratio, and $\alpha_{\text{new}}$ represents the policy state post-update as modified by the stochastic dropout mask. This structural stochasticity, compounded with per-step policy updates, severely undermines training convergence. The training hyperparameters are identical to those used for $\pi_0$, with Flow-SDE employed as the primary RL algorithm.

\subsection{Results}
Results are summarized in \cref{tab:gr00t_result}. For the few-shot model, the SFT baseline achieves only a 52.5\% success rate, reflecting limited generalization from sparse demonstrations. Conversely, our RL-based Flow-SDE significantly improves performance to 89.9\%. These results, obtained using default $\pi_0$ configurations, underscore the broad applicability of our method across architectures. While task-specific tuning could further enhance performance, we leave such optimization for future work.

\section{Limitations and Future Work}
\label{appendix:limiation}
\textbf{Noise Injection.}
Our current noise injection strategy exhibits a performance drop during the ODE-to-SDE conversion. Flow-CPS \citep{wang2025coefficients} attributes this loss to numerical error and proposes an improved coefficients-preserving sampling method. In our experiments, we attempted this configuration. Consistent with our hyperparameter ablation, our experiments showed that while this configuration mitigated the ODE-SDE precision error, it yielded limited RL improvement. Nevertheless, we argue that improving the noise injection strategy holds significant potential, specifically converting the ODE formulation to an SDE formulation while preserving the action distribution undisturbed.

\textbf{Training Acceleration.}
Our current implementation of the mixed ODE-SDE rollout is simplistic in Flow-SDE, \ie, it randomly selects one denoising step as an SDE step, while all other steps remain ODE steps. We posit that future investigations into mixed ODE-SDE rollouts, leveraging advances in accelerating flow-based image generation \citep{li2025mixgrpo, he2025tempflow, liu2025flowgrpo, li2025branchgrpo}, could further enhance Flow-SDE, leading to faster training and improved performance.

\textbf{Generalization.}
Maniskill OOD tests indicate that the semantic generalization of SFT and RL models remains limited. To address this, future work will leverage RL to enhance robustness by training on more diverse task distributions and varied linguistic instructions, thereby fostering better cross-task adaptability.

\section{Training Hyperparameters.}
\label{appendix:hyper_params}
We record the training hyperparameters used to train both $\pi_{0}$ and $\pi_{0.5}$ on each benchmark, and present them in \cref{tab:hyperparam_table,tab:hyperparam_table2,tab:hyperparam_table3}.

\begin{table*}[htb]
\centering
\caption{Hyperparameters across LIBERO.}
\label{tab:hyperparam_table}
\footnotesize 
\setlength{\tabcolsep}{0pt} 
\begin{tabular*}{\linewidth}{@{\extracolsep{\fill}}lcccccccc}
\toprule
\multirow{4}{*}{\textbf{Parameters}} 
  & \multicolumn{8}{c}{\textbf{Algorithms and Tasks}} \\
\cmidrule(lr){2-9}
 & \multicolumn{4}{c}{$\boldsymbol{\pi_0}$} 
 & \multicolumn{4}{c}{$\boldsymbol{\pi_{0.5}}$} \\
\cmidrule(lr){2-5} \cmidrule(lr){6-9}
 & \textbf{Spatial} & \textbf{Object} & \textbf{Goal} & \textbf{Long}
 & \textbf{Spatial} & \textbf{Object} & \textbf{Goal} & \textbf{Long} \\
\midrule
Train epochs                   & 500  & 500  & 500  & 500  & 500  & 500  & 500  & 500  \\
Global batch size              & 2048 & 2048 & 2048 & 2048 & 2048 & 2048 & 2048 & 2048 \\
Update epochs                  & 4    & 4    & 4    & 4    & 1    & 1    & 3    & 4    \\
Actor lr                       & 5e-6 & 5e-6 & 5e-6 & 5e-6 & 5e-6 & 5e-6 & 5e-6 & 5e-6 \\
Critic lr                      & 1e-4 & 1e-4 & 1e-4 & 1e-4 & 1e-4 & 1e-4 & 1e-4 & 1e-4 \\
Scheduler                      & False & False & False & False & False & False & False & True \\
\midrule
Reward discount rate $\gamma$  & 0.99 & 0.99 & 0.99 & 0.99 & 0.99 & 0.99 & 0.99 & 0.99 \\
GAE $\lambda$                  & 0.95 & 0.95 & 0.95 & 0.95 & 0.95 & 0.95 & 0.95 & 0.95 \\
Clip ratio $\epsilon$          & 0.2  & 0.2  & 0.2  & 0.2  & 0.2  & 0.2  & 0.2  & 0.2  \\
\midrule
Interaction steps              & 240  & 240  & 320  & 480  & 240  & 320  & 320  & 480  \\
Parallel environments          & 64   & 64   & 64   & 64   & 64   & 64   & 64   & 64   \\
Rollout epochs                 & 8    & 8    & 8    & 8    & 8    & 8    & 8    & 8    \\
\midrule
Action prediction horizon $H$  & 50    & 50    & 50    & 50   & 10    & 10    & 10    & 10   \\
Action replan horizon $H'$  & 5    & 5    & 5    & 10   & 5    & 5    & 5    & 10   \\
Denoise steps                  & 4    & 4    & 4    & 4    & 3    & 5    & 5    & 5    \\
Noise level $\sigma$ (Flow-SDE)& 0.5  & 0.5  & 0.5  & 0.5  & 0.5  & 0.3  & 0.3  & 0.5  \\
Max log-var (Flow-Noise)       & 0.16 & 0.16 & 0.16 & 0.16 & 0.10 & 0.10 & 0.10 & 0.10 \\
Min log-var (Flow-Noise)       & 0.08 & 0.08 & 0.08 & 0.08 & 0.04 & 0.04 & 0.04 & 0.04 \\
Entropy bonus (Flow-Noise)     & 0.005 & 0.005 & 0.005 & 0.005 & 0.005 & 0.005 & 0.005 & 0.005 \\
\bottomrule
\end{tabular*}
\end{table*}

\begin{table*}[htb]
\centering
\caption{Hyperparameters across SIMPLER and ManiSkill.}
\label{tab:hyperparam_table2}
\footnotesize 
\setlength{\tabcolsep}{0pt}
\begin{tabular*}{\linewidth}{@{\extracolsep{\fill}}lcccccccccc}
\toprule
\multirow{4}{*}{\textbf{Parameters}} 
  & \multicolumn{10}{c}{\textbf{Algorithms and Tasks}} \\
\cmidrule(lr){2-11}
 & \multicolumn{5}{c}{$\boldsymbol{\pi_0}$} 
 & \multicolumn{5}{c}{$\boldsymbol{\pi_{0.5}}$} \\
\cmidrule(lr){2-6} \cmidrule(lr){7-11}
 & \textbf{Eggplant} & \textbf{Carrot} & \textbf{Spoon} & \textbf{Cube} & \textbf{ManiSkill}
 & \textbf{Eggplant} & \textbf{Carrot} & \textbf{Spoon} & \textbf{Cube} & \textbf{ManiSkill} \\
\midrule
SFT train steps                & 1000 & 1000 & 1000 & 1000 & 1000 & 1000 & 1000 & 1000 & 1000 & 1000 \\
RL train steps                 & 40   & 40   & 40   & 130  & 150  & 40   & 40   & 40   & 70   & 150  \\
Global batch size              & 2560 & 2560 & 2560 & 2560 & 5120 & 2560 & 2560 & 2560 & 2560 & 5120 \\
Update epochs                  & 4    & 4    & 4    & 4    & 4    & 4    & 4    & 4    & 4    & 5    \\
Actor lr                       & 5.6e-6 & 5.6e-6 & 5.6e-6 & 5.6e-6 & 7.91e-6 & 5.6e-6 & 5.6e-6 & 5.6e-6 & 5.6e-6 & 7.91e-6 \\
Critic lr                      & 1.1e-4 & 1.1e-4 & 1.1e-4 & 1.1e-4 & 1.55e-4 & 1.1e-4 & 1.1e-4 & 1.1e-4 & 1.1e-4 & 1.55e-4 \\
Scheduler                      & False & False & False & False & False & False & False & False & False & False \\
\midrule
Reward discount rate $\gamma$  & 0.99 & 0.99 & 0.99 & 0.99 & 0.99 & 0.99 & 0.99 & 0.99 & 0.99 & 0.99 \\
GAE $\lambda$                  & 0.95 & 0.95 & 0.95 & 0.95 & 0.95 & 0.95 & 0.95 & 0.95 & 0.95 & 0.95 \\
Clip ratio $\epsilon$          & 0.2  & 0.2  & 0.2  & 0.2  & 0.2  & 0.2  & 0.2  & 0.2  & 0.2  & 0.2  \\
\midrule
Interaction steps              & 48   & 48   & 48   & 48   & 48   & 48   & 48   & 48   & 48   & 48   \\
Parallel environments          & 256  & 256  & 256  & 256  & 320  & 256  & 256  & 256  & 256  & 320  \\
Rollout epochs                 & 1    & 1    & 1    & 1    & 1    & 1    & 1    & 1    & 1    & 1    \\
\midrule
Action prediction horizon $H$  & 8    & 8    & 8    & 8    & 8    & 8    & 8    & 8    & 8    & 8    \\
Action replan horizon $H^\prime$ & 5   & 5    & 5    & 5    & 5    & 5    & 5    & 5    & 5    & 5    \\
Denoise steps                  & 4    & 4    & 4    & 4    & 4    & 4    & 4    & 4    & 4    & 4    \\
Noise level $\sigma$ (Flow-SDE)& 0.5  & 0.5  & 0.5  & 0.5  & 0.5  & 0.5  & 0.5  & 0.5  & 0.5  & 0.5  \\
Max log-var (Flow-Noise)       & 0.16 & 0.16 & 0.16 & 0.16 & 0.16 & 0.10 & 0.10 & 0.10 & 0.10 & 0.10 \\
Min log-var (Flow-Noise)       & 0.08 & 0.08 & 0.08 & 0.08 & 0.08 & 0.04 & 0.04 & 0.04 & 0.04 & 0.04 \\
Entropy bonus (Flow-Noise)     & 0.005 & 0.005 & 0.005 & 0.005 & 0.005 & 0.005 & 0.005 & 0.005 & 0.005 & 0.005 \\
\bottomrule
\end{tabular*}
\end{table*}

\begin{table*}[htb]
\centering
\caption{Hyperparameters across MetaWorld and CALVIN benchmarks.}
\label{tab:hyperparam_table3}
\footnotesize 
\setlength{\tabcolsep}{0pt}
\begin{tabular*}{\linewidth}{@{\extracolsep{\fill}}lcccc}
\toprule
\multirow{4}{*}{\textbf{Parameters}} 
  & \multicolumn{4}{c}{\textbf{Benchmarks and models}} \\
\cmidrule(lr){2-5}
 & \multicolumn{2}{c}{\textbf{MetaWorld}} 
 & \multicolumn{2}{c}{\textbf{CALVIN}} \\
\cmidrule(lr){2-3} \cmidrule(lr){4-5}
 & $\boldsymbol{\pi_0}$ & $\boldsymbol{\pi_{0.5}}$ & $\boldsymbol{\pi_0}$ & $\boldsymbol{\pi_{0.5}}$ \\
\midrule
Train epochs                   & 450 & 450 & 100 & 100 \\
Global batch size              & 2048 & 2048 & 2048 & 2048 \\
Update epochs                  & 4    & 4    & 4    & 4    \\
Actor lr                       & 1e-5 & 5e-6 & 5e-6 & 5e-6 \\
Critic lr                      & 1e-4 & 1e-4 & 1e-4 & 1e-4 \\
Scheduler                      & False & True & False & False \\
\midrule
Reward discount rate $\gamma$  & 0.99 & 0.99 & 0.99 & 0.99 \\
GAE $\lambda$                  & 0.95 & 0.95 & 0.95 & 0.95 \\
Clip ratio $\epsilon$          & 0.2  & 0.2  & 0.2  & 0.2  \\
\midrule
Interaction steps              & 100  & 100  & 480  & 480  \\
Parallel environments          & 64   & 64   & 64   & 64   \\
Rollout epochs                 & 8    & 8    & 8    & 8    \\
\midrule
Action prediction horizon $H$  & 5    & 5    & 5    & 5    \\
Action replan horizon $H'$  & 5    & 5    & 5    & 5    \\
Denoise steps                  & 5    & 5    & 5    & 5    \\
Noise level $\sigma$ (Flow-SDE)& 0.5  & 0.5  & 0.5  & 0.5  \\
Max log-var (Flow-Noise)       & 0.10 & 0.10 & 0.16 & 0.16 \\
Min log-var (Flow-Noise)       & 0.04 & 0.04 & 0.08 & 0.08 \\
Entropy bonus (Flow-Noise)     & 0.005 & 0.005 & 0.005 & 0.005 \\
\bottomrule
\end{tabular*}
\end{table*}

\end{document}